\documentclass[final,5p,times]{elsarticle}

\journal{Knowledge-Based Systems}

\usepackage{amsmath,amssymb}
\usepackage{graphicx}
\usepackage{booktabs}
\usepackage{array}
\usepackage{makecell}
\usepackage{multirow}
\usepackage{float}
\usepackage{stfloats}
\usepackage{placeins}
\usepackage{cuted}
\usepackage{capt-of}
\usepackage{algpseudocode}
\usepackage{hyperref}
\usepackage{enumitem}

\hypersetup{
 colorlinks=true,
 linkcolor=blue,
 citecolor=blue,
 urlcolor=blue
}

\biboptions{sort}

\newcommand{\model}{\textsc{TCHG}}
\newcommand{\bestval}[1]{\textbf{#1}}

\floatstyle{ruled}
\newfloat{algorithm}{tbp}{loa}
\floatname{algorithm}{Algorithm}

\begin{document}

\begin{frontmatter}

\title{TCHG: Tri-Trust Conditioned Heterogeneous Graph Learning for Reliable Dynamic Trust Prediction}

\author[inst1]{Bohao Liao}
\ead{liaobohao@stu.xidian.edu.cn}
\author[inst2]{Boyu Deng}
\ead{kdydby2014@sina.com}
\author[inst3]{Qipeng Song}
\ead{qpsong@xidian.edu.cn}
\author[inst4]{Jieling Wang}
\ead{jlwang@xidian.edu.cn}
\author[inst5]{Jingchao Wang\corref{cor1}}
\ead{wangjc.2000@tsinghua.org.cn}
\cortext[cor1]{Corresponding authors.}
\affiliation[inst1]{
 organization={School of Telecommunication Engineering, Xidian University},
 city={Xi'an},
 postcode={710071},
 state={Shaanxi},
 country={China}
}
\affiliation[inst2]{
 organization={National Key Laboratory of Multi-domain Data Collaborative Processing and Control},
 city={Beijing},
 postcode={100141},
 country={China}
}
\affiliation[inst3]{
 organization={School of Cyber Engineering, Xidian University},
 city={Xi'an},
 postcode={710071},
 state={Shaanxi},
 country={China}
}
\affiliation[inst4]{
 organization={School of Telecommunication Engineering, Xidian University},
 city={Xi'an},
 postcode={710071},
 state={Shaanxi},
 country={China}
}
\affiliation[inst5]{
 organization={National Key Laboratory of Multi-domain Data Collaborative Processing and Control},
 city={Beijing},
 postcode={100141},
 country={China}
}

\begin{abstract}
Trust prediction infers latent user–user trust relations and provides important support for social recommendation, fake-review and manipulation detection, and risk identification. Graph neural networks have become a prominent approach to trust prediction because of their ability to learn network structures and complex trust dependencies. However, existing methods often rely on a unified representation of trust signals and do not disentangle heterogeneous trust evidence into separate evidence channels, failing to exploit the distinct roles that different evidence channels should play during trust modeling.

To address this gap, this paper argues that trust evidence should not be treated as an undifferentiated input, but should be decomposed and used as functional control factors over graph propagation. We propose TCHG, a tri-trust conditioned heterogeneous graph learning framework that decomposes trust evidence into three channels and assigns them distinct functional roles in propagation: entity reliability governs message admission, interaction-behavior reliability modulates propagation strength, and contextual trust adjusts the propagation mode through context-conditioned operator selection. Since the three evidence channels evolve at different temporal scales, TCHG maintains independent temporal states with non-uniform decay rates to prevent rapidly changing contextual signals from overwriting slowly accumulated entity reliability. It further predicts trust probability and calibrates the output probability, improving predictive confidence under sparse or conflicting evidence. Extensive experiments on multiple public trust datasets show that TCHG achieves effective and reliable trust prediction compared with representative trust prediction and heterogeneous graph baselines.

\end{abstract}


\end{frontmatter}

\section{Introduction}

Trust prediction aims to infer whether one user is likely to trust another. On online platforms, users rate items, write reviews, and form trust or following relations, and these relations shape which content is recommended and which users are treated as reliable information sources~\cite{sherchan2013,jiang2016,ghafari2020}. Trust is commonly understood as a relational judgment involving the trustor's willingness to rely on another party under uncertainty~\cite{mayer1995,mcknight2001}. Trust prediction therefore underpins a range of Web applications, including social recommendation, online review moderation, content filtering, and risk identification. The stakes are high: once an unreliable user is mistakenly predicted as trustworthy, low-quality reviews or suspicious activity can spread through the user network and contaminate downstream decisions. This risk continues to grow as fake reviews, collusive behavior, and AI-generated content become ever easier to produce~\cite{ftc2024,shepardson2024,meng2025}.

Graph neural networks (GNNs) provide a natural modeling paradigm for trust prediction. Trust data can be organized as a graph that combines explicit user--user trust links, user--item ratings or reviews, relation types, and contextual information. Through message passing and neighborhood aggregation, GNNs capture structural signals from local and higher-order neighborhoods, which matters because trust relations are typically transitive, composable, and directional. Compared with matrix factorization, similarity-based methods, and rule-based reasoning, GNNs also reduce the reliance on hand-crafted rules and shallow feature engineering by learning node representations, relation patterns, and neighborhood structures in an end-to-end manner~\cite{kipf2017,hamilton2017,velickovic2018}. These properties have made GNNs a mainstream technical direction for trust prediction.

\begin{figure}[!t]
\centering
\includegraphics[width=\columnwidth]{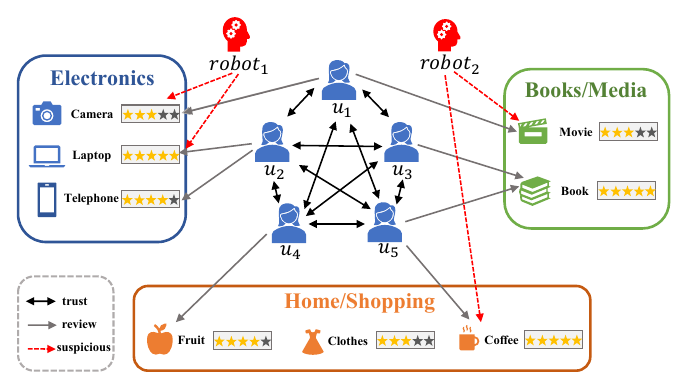}
\caption{A typical Web trust network. Predicting whether one user trusts another depends on three types of evidence with distinct roles: the reliability of the information source (user entities), the normality of interaction behavior, and the contextual conditions of the current scenario.}
\label{fig:scenario}
\end{figure}

Along this direction, existing studies have explored structural propagation, heterogeneous information fusion, temporal modeling, context awareness, and robustness enhancement~\cite{lin2020,huo2024,yu2023,wang2023trustguard,lin2021medley,wang2026cat}, and have substantially advanced the field. They nevertheless share a common limitation in \emph{how trust evidence is used}: heterogeneous evidence about user entities, interaction behavior, and contextual conditions is typically treated as undifferentiated input, concatenated into node features, compressed into attention weights, or absorbed into path embeddings, and then consumed by an essentially uniform aggregation process. Figure~\ref{fig:scenario} illustrates why this is restrictive. In a typical Web trust network, judging whether one user should trust another involves three questions of different natures: \emph{Is the information source itself reliable} (an established user, or a bot-like account)? \emph{Is the current interaction behavior normal} (a stable co-rating history, or bursty low-rating activity)? \emph{How does the current scenario condition the trust judgment} (trust built on electronics purchases may not transfer to book reviews)? Answering these questions requires the corresponding evidence to play different functional roles during propagation: deciding whether a message should be admitted at all, modulating how strongly an admitted message propagates, and selecting the mode in which it propagates. Once all evidence is flattened into a single representation or importance weight, these distinctions are lost. For example, the model can no longer reject messages from a suspicious source outright while merely down-weighting messages tied to mildly anomalous behavior.

Closing this gap calls for a shift in perspective: heterogeneous trust evidence should serve not as representation enhancement but as \emph{propagation control}. Pursuing this evidence-controlled route, however, raises three technical challenges.

\textbf{TC1: How can decomposed trust evidence be turned into selective propagation control?} Neighborhood aggregation, attention mechanisms, and meta-path modeling all learn the relative importance of neighbors, relations, or paths. Importance weighting, however, offers only a single degree of freedom: it cannot simultaneously express whether a source message should be admitted, how strongly an admitted message should propagate, and whether the propagation mode should switch with the scenario. The challenge is to design a propagation mechanism in which entity, behavior, and context evidence each exert a distinct, well-defined control function, while keeping the model end-to-end trainable.

\textbf{TC2: How should evidence channels with different temporal stability be memorized?} Decomposing trust evidence into separate channels immediately exposes their asynchronous dynamics: long-term user reliability accumulates slowly, interaction behavior mixes long-term patterns with recent changes, and contextual conditions fluctuate rapidly with time, relations, and local scenarios. A single shared memory updated at a uniform rate either lets volatile contextual signals overwrite slowly accumulated reliability, or dulls the model's sensitivity to recent behavioral anomalies. The challenge is to retain, update, and decay each type of historical information at its own temporal scale.

\textbf{TC3: How can probability outputs remain reliable under sparse or conflicting evidence?} Trust prediction is not only a candidate-edge ranking task; its outputs often feed risk-sensitive decisions such as risk identification, suspicious-behavior screening, and relation-level filtering. Ranking and discrimination metrics such as MRR, AP, and AUC, however, do not reveal whether a model is overconfident on high-risk samples. When user histories are sparse, candidate relations lack sufficient evidence, or different evidence channels conflict, the model should report not only a trust probability but also the risk attached to that prediction.

To address these challenges, we propose \model{} (Tri-Trust Conditioned Heterogeneous Graph Learning), an evidence-controlled heterogeneous graph learning framework for reliable dynamic trust prediction. \model{} first constructs structured tri-trust evidence, comprising entity reliability, interaction-behavior reliability, and contextual trust, from historical trust links, user--item interactions, temporal statistics, and local graph structure. The three channels are then assigned distinct functional roles in propagation: entity reliability gates message admission, behavior reliability modulates propagation strength, and contextual trust selects the propagation operator. To respect their asynchronous dynamics, \model{} maintains component-decoupled temporal memories with non-uniform decay rates and masked updates. Finally, \model{} predicts sample-level uncertainty alongside trust probability and calibrates its probability outputs, aligning predictive confidence with actual error risk. In this way, evidence construction, dynamic graph learning, and reliable prediction are connected within a unified framework.

We evaluate \model{} on three public trust datasets: Epinions, Ciao, and CiaoDVD. Epinions provides native trust timestamps and is used for dynamic trust prediction, with separate evaluation of observed-user and unobserved-user scenarios; Ciao and CiaoDVD lack native trust timestamps and are therefore evaluated under all-users random trust-link prediction splits, without constructing pseudo trust timestamps. Compared with representative trust prediction models and heterogeneous graph baselines, \model{} achieves consistent gains on the main metrics. Under the 80\%-10\%-10\% split on Epinions, it reaches MRR scores of 0.8811 and 0.7018 in the observed-user and unobserved-user scenarios, respectively. Under the same split, it obtains MRR scores of 0.6499 and 0.9325 on Ciao and CiaoDVD, respectively. Ablation studies, temporal-memory analysis, efficiency analysis, reliability evaluation, robustness evaluation, and selective prediction further validate the effectiveness of the proposed designs.

The main contributions of this paper are summarized as follows.

\begin{itemize}[leftmargin=2em]
\item First, we propose tri-trust evidence construction together with an evidence-controlled heterogeneous propagation mechanism. \model{} builds three evidence channels, namely entity reliability, interaction-behavior reliability, and contextual trust, from users' historical states, interaction patterns, and contextual conditions, and maps them to message admission, propagation-strength modulation, and context-conditioned operator selection, respectively. Heterogeneous trust evidence thereby directly constrains graph propagation, rather than serving only as additional input features.

\item Second, we design a component-decoupled temporal trust memory mechanism. Separate memory states, non-uniform temporal decay, and masked updates allow the model to track entity, behavior, and context evidence at their own temporal scales, preventing long-term status, recent behavior, and contextual fluctuations from being conflated in a shared memory state.

\item Third, we introduce uncertainty-aware output modeling for reliable decision making. The model estimates sample-level uncertainty together with trust probability, and combines posterior probability calibration with alignment between prediction error and uncertainty. This improves the consistency between predictive confidence and actual error risk, and provides auxiliary reliability signals for downstream risk-sensitive applications.
\end{itemize}

\section{Related Work}

This section reviews four lines of work related to \model{}: GNN-based trust prediction, dynamic heterogeneous graph learning, context-aware trust prediction, and uncertainty estimation with probability calibration.

\subsection{GNN-Based Trust Prediction}

Trust prediction has been widely studied in online social networks, recommender systems, and trusted computing. Early studies inferred latent trust relations through social regularization, random walks, transitive trust inference, subjective logic, and time-aware matrix factorization~\cite{massa2007,ma2009,jamali2009,yao2013,liu2014,liu2017,gao2021,forsati2014}. With the development of graph neural networks, trust prediction has increasingly been formulated as a graph representation learning problem, where user representations, trust links, and neighborhood structures are learned in an end-to-end manner.

Recent GNN-based trust models have improved trust prediction from different perspectives, including higher-order neighborhood modeling, multi-aspect user attention, trust propagation and composability, knowledge-enhanced trust semantics, robustness, and temporal variation~\cite{lin2020,jiang2022,huo2024,yu2023,wang2023trustguard,lin2021medley}. These studies demonstrate the value of structural, attribute, semantic, and temporal information for trust modeling. However, such information is usually used as input features, attention weights, path representations, or unified propagation states. They do not explicitly assign different trust evidence channels to separate propagation functions, such as source-message admission, behavior-aware strength modulation, and context-conditioned propagation-mode selection. This functional distinction is the main difference between existing GNN-based trust models and \model{}.

\subsection{Dynamic Heterogeneous Graph Learning}

Trust networks are naturally heterogeneous and dynamic because users form trust relations, interact with items, and generate temporal behavior records. Heterogeneous graph neural networks provide important tools for modeling typed nodes and relations. RGCN learns relation-specific transformations, HAN captures meta-path semantics through hierarchical attention, HGT introduces type-aware attention over node and edge types, and heterogeneous attention models further support structural learning across typed graph components~\cite{schlichtkrull2018,wang2019han,hu2020,hong2020}. Dynamic graph methods further model temporal interactions through temporal encoding, recurrent memory, or temporal attention, as seen in JODIE, TGAT, TGN, and DySAT~\cite{kumar2019,xu2020,rossi2020,sankar2020}.

These models provide general graph-learning foundations, but their objectives are not specifically designed for trust evidence modeling. They usually learn relation-level transformations, semantic attention, or unified temporal states, while trust prediction requires a finer distinction among slowly accumulated entity reliability, behavior evidence that may change with recent interactions, and contextual conditions that fluctuate across time and scenarios. \model{} builds on dynamic heterogeneous graph learning, but further imposes functional and temporal-scale constraints on trust evidence.

\subsection{Context-Aware Trust Prediction}

Context-aware trust prediction emphasizes that trust is affected not only by user--user trust links, but also by user behavior, item interactions, temporal factors, and local scenarios. Related studies enhance user representations through user--item interactions, item categories, contextual paths, and temporal information, building on meta-path similarity and representation learning for heterogeneous information networks~\cite{sun2011,dong2017}. Among them, CAT is closest to our task. It combines dynamics, heterogeneity, and context through context-aware meta-paths, temporal encoding, recent temporal-neighbor sampling, dual attention, one-hop trust propagation, and context aggregation~\cite{wang2026cat}.

CAT provides an important reference because it demonstrates the value of contextual information in trust modeling. However, its context mainly affects path representations and aggregation weights. It does not separately model whether a source message should be admitted, how strongly an admitted message should propagate, or which propagation operator should be used under the current scenario. \model{} addresses this boundary by mapping entity reliability, interaction-behavior reliability, and contextual trust to entity gating, behavior modulation, and context-conditioned operator selection, respectively.

\subsection{Uncertainty Estimation and Probability Calibration}

Neural networks often produce poorly calibrated outputs, where high confidence does not necessarily correspond to high empirical correctness~\cite{guo2017}. This issue is particularly relevant to trust prediction because trust scores may support social recommendation, suspicious-behavior screening, relation filtering, and other risk-sensitive decisions. Therefore, trust prediction should be evaluated not only by ranking or discrimination quality, but also by the reliability of its predicted probabilities.

Uncertainty estimation and probability calibration provide useful tools for this purpose. Calibration metrics such as ECE, Brier Score, and NLL measure whether predicted probabilities match observed outcomes, while methods such as posterior calibration, Bayesian learning, and deep ensembles aim to improve or estimate predictive reliability~\cite{brier1950,gneiting2007,naeini2015,niculescu2005,lakshminarayanan2017,kendall2017,platt1999,kull2019}. Existing graph-based trust prediction studies, including CAT and KGTrust, mainly report metrics such as MRR, AP, and AUC, with limited attention to whether prediction confidence matches actual error risk. \model{} therefore introduces uncertainty-aware prediction and posterior probability calibration, treating probability reliability as a complementary objective to ranking performance.

\section{Problem Formulation}

This section formalizes the dynamic user trust prediction problem. We first define the dynamic heterogeneous trust graph and the event-based trust relation, then decompose the evidence of each candidate event into tri-trust evidence, specify the historical-availability settings used for evaluation, and finally state the prediction objective.

Unlike general link prediction~\cite{liben2007,clauset2008,adamic2003}, user trust relations are shaped not only by historical structure but also by user entity states, interaction patterns, and contextual conditions. We therefore define the task as dynamic heterogeneous graph prediction driven by entity, behavior, and context evidence, while following standard representations from heterogeneous relation modeling~\cite{schlichtkrull2018,wang2019han,hu2020} and dynamic graph event modeling~\cite{kumar2019,xu2020,rossi2020}.

\textbf{Definition 1. Dynamic heterogeneous trust graph.}
Given the user set $\mathcal{U}$, the relation type set $\mathcal{R}$ and the time interval $[0,T]$, the dynamic heterogeneous trust graph is expressed as
$\mathcal{G}(T)=(\mathcal{V}(T),\mathcal{E}(T),\mathcal{R},\mathcal{X},\Phi,\Psi)$,
where $\mathcal{V}(T)\subseteq\mathcal{U}$ is the set of user nodes that appeared within time $T$, and $\mathcal{E}(T)$ is the set of observed timestamped relation edges before $T$.
$\Phi:\mathcal{V}\rightarrow\mathcal{A}$ and
$\Psi:\mathcal{E}\rightarrow\mathcal{R}$ denote the node-type and edge-type mapping functions respectively, and
$\mathcal{X}$ denotes the set of evidence features associated with nodes, edges, and events.
When multiple relation types or multi-source evidence signals exist in the graph, the graph is regarded as a heterogeneous trust graph.

\textbf{Definition 2. Event-based trust relationship.}
The dynamic trust network is represented by a time-ordered event stream
$\mathcal{S}=\{e_1,e_2,\ldots,e_{|\mathcal{S}|}\}$, where a candidate trust event is defined as
$e_{ij,t}=(u_i,u_j,r,t,y_{ij,t},\mathbf{x}_{ij,t})$.
Here $u_i,u_j\in\mathcal{U}$ are the trust initiator and trust receiver,
$r\in\mathcal{R}$ is the relation type, $t$ is the event time,
$y_{ij,t}\in\{0,1\}$ is the ground-truth trust label, and
$\mathbf{x}_{ij,t}$ denotes the evidence features of this candidate event.
At prediction time $t$, the model has access only to the historical graph $\mathcal{G}_{<t}$; that is, all information used for feature construction and inference originates from events before the target event.

\textbf{Definition 3. Tri-trust evidence.}
To characterize how trust is formed, we decompose the evidence features of a candidate event $e_{ij,t}$ into three complementary types, $\mathbf{x}_{ij,t}^{E}$, $\mathbf{x}_{ij,t}^{B}$ and $\mathbf{x}_{ij,t}^{C}$, consistent with the view that user behavior, temporal factors, and local scenarios jointly affect trust relations~\cite{wang2026cat}:
$\mathbf{x}_{ij,t}^{E}$ denotes entity reliability evidence describing the long-term trust status of the two endpoints;
$\mathbf{x}_{ij,t}^{B}$ denotes interaction-behavior reliability evidence describing whether the local interaction is stable, frequent, or abnormal; and
$\mathbf{x}_{ij,t}^{C}$ denotes contextual trust evidence describing the temporal, relational, and local-scenario conditions of the current event.
The three types respectively answer ``Who is trustworthy?'', ``Is the interaction reliable?'', and ``How should trust information be disseminated in the current scenario?''
The input evidence for a candidate event is thus written as
$\mathbf{x}_{ij,t}=\{\mathbf{x}_{ij,t}^{q}\mid q\in\{E,B,C\}\}$.

\textbf{Definition 4. Observed-user and unobserved-user settings.}
To evaluate prediction under different historical-information conditions, we distinguish two test scenarios.
A sample belongs to the \emph{observed-user} scenario if both endpoints of the candidate pair appear in the training graph and have available historical trust or interaction evidence before the target prediction time.
A sample belongs to the \emph{unobserved-user} scenario if at least one endpoint does not appear in the training graph, or lacks available historical evidence before the target prediction time.
This division characterizes historical availability for datasets with native trust timestamps, such as Epinions, and should be distinguished from the observation mask $o_t$ in the temporal memory module: the former is a task-scenario definition, whereas the latter is an internal variable that controls whether the current event is written into the memory state. For Ciao and CiaoDVD, which lack native trust timestamps, this paper uses all-users random trust-link prediction splits and does not apply this observed/unobserved subdivision.

\textbf{Task Definition.}
Given a candidate user pair $(u_i,u_j)$, relation type $r$, target time $t$, historical dynamic graph $\mathcal{G}_{<t}$, and tri-trust evidence
$\{\mathbf{x}_{ij,t}^{E},\mathbf{x}_{ij,t}^{B},\mathbf{x}_{ij,t}^{C}\}$,
dynamic user trust prediction aims to learn a parameterized function $f_{\Theta}(\cdot)$:
\begin{equation}
 (\hat{y}_{ij,t},\hat{u}_{ij,t})
 =
 f_{\Theta}
 \big(
 u_i,u_j,r,t,\mathcal{G}_{<t},
 \mathbf{x}_{ij,t}^{E},
 \mathbf{x}_{ij,t}^{B},
 \mathbf{x}_{ij,t}^{C}
 \big),
\end{equation}
where $\hat{y}_{ij,t}\in[0,1]$ is the predicted probability that user $u_i$ trusts user $u_j$, and $\hat{u}_{ij,t}\in[0,1]$ is the model's uncertainty estimate for this prediction.
In summary, the problem is: how can entity reliability, interaction-behavior reliability, and contextual trust be converted into effective graph learning signals, so that trust relations can be predicted accurately and reliably under native-timestamp dynamic settings, and remain effective under all-users random trust-link prediction splits when trust timestamps are unavailable?

\section{The \model{} Framework}

This section first presents the overall workflow of \model{}, and then defines tri-trust evidence representation, conditioned propagation, component memory, and uncertainty-aware calibration.

\subsection{Overall Framework}

\model{} contains four steps: tri-trust evidence construction, conditioned propagation, component memory, and reliable output. Given candidate trust events, \model{} builds entity, behavior, and context evidence from the available historical graph or training graph, and outputs trust probability and predictive uncertainty. For datasets with native trust timestamps, prediction strictly uses the history before the target event; for datasets without native trust timestamps, evidence is constructed from the training graph and available interaction records under all-users random trust-link prediction splits. The three evidence channels use independent encoders and memory states, and play different control roles in propagation.

For candidate events $e_{ij,t}$, the overall prediction process of \model{} can be expressed as:
\begin{equation}
 \label{eq:tchg-overall}
 \hat{y}_{ij,t},\hat{u}_{ij,t}
 =
 \model(
 u_i,u_j,r,t,
 \mathbf{x}_{ij,t}^{E},
 \mathbf{x}_{ij,t}^{B},
 \mathbf{x}_{ij,t}^{C},
 \mathbf{e}_{ij,t},
 \mathcal{G}_{<t}
 ),
\end{equation}
where $\hat{y}_{ij,t}$ is the predicted probability that user $u_i$ trusts user $u_j$, $\hat{u}_{ij,t}$ is the model's uncertainty estimate, $\mathbf{e}_{ij,t}$ denotes edge-level auxiliary features, and $\mathcal{G}_{<t}$ denotes the historical dynamic graph available before the target time $t$.

\begin{figure*}[!t]
 \centering
 \includegraphics[width=\textwidth,trim=55 90 95 95,clip]{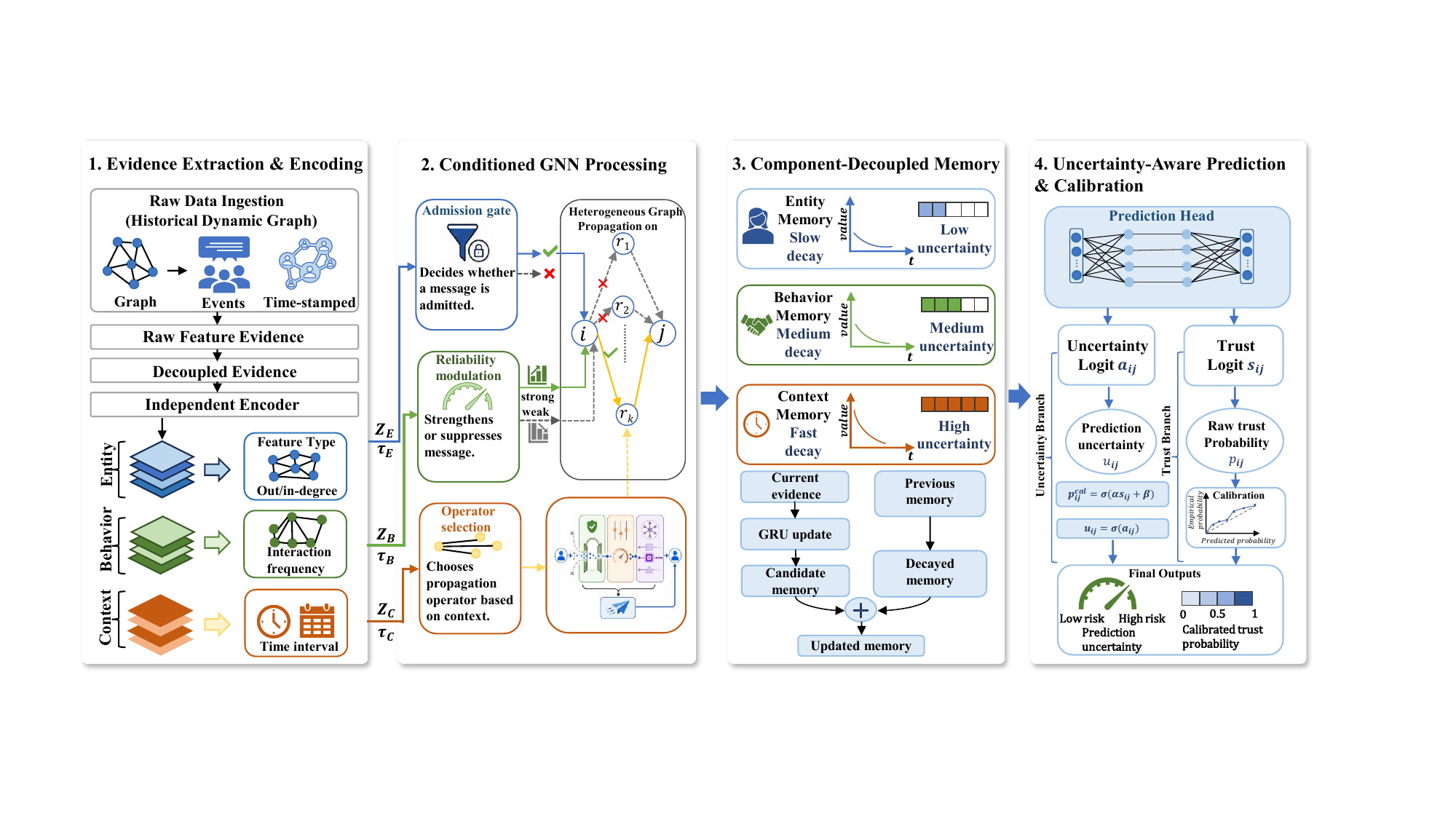}
 \caption{The overall structure of \model{} includes tri-trust evidence construction and encoding, tri-trust conditioned heterogeneous propagation, component-decoupled temporal trust memory, and uncertainty-aware prediction and calibration.}
 \label{fig:algorithm}
\end{figure*}

As shown in Figure~\ref{fig:algorithm}, \model{} sequentially performs online evidence construction and encoding, tri-trust conditioned heterogeneous propagation, component-level memory reading and updating, and uncertainty-aware prediction and calibration. When predicting the current event, the model reads only the graph and memory available before the target time; the current event becomes part of the history only after prediction is completed and memory writing is allowed.

Table~\ref{tab:notation} summarizes the main symbols.

\begin{table}[!t]
 \centering
 \scriptsize
 \caption{Main symbols and their meanings.}
 \label{tab:notation}
 \begin{tabular}{@{}>{\raggedright\arraybackslash}p{0.32\linewidth}>{\raggedright\arraybackslash}p{0.61\linewidth}@{}}
 \toprule
 Symbol & Meaning\\
 \midrule
 $e_{ij,t}=(u_i,u_j,r,t)$ & Candidate trust event from trust-initiating user $u_i$ to trust-receiving user $u_j$. \\
 $\mathcal{G}_{<t}$ & Historical dynamic graph before target time $t$. \\
 $\{\mathbf{x}_{ij,t}^{q}\}_{q\in\{E,B,C\}}$ & Entity, behavior, and context evidence channels of candidate event $e_{ij,t}$. \\
 $(\mathbf{z}_{ij,t}^{q},\tau_{ij,t}^{q})$ & Latent representation and credibility strength of evidence channel $q$. \\
 $\tau_{ij,t}^{E}$ & Pair-level entity reliability strength of candidate relation $(u_i,u_j,t)$. \\
 $\tau_E^{\mathrm{src}}(v,t)$ & Source-side entity reliability strength of incoming source user $v$ before time $t$; used only for message admission. \\
 $\{g_E,\rho_B,\boldsymbol{\alpha}_C\}$ & Entity message-admission gate, behavior modulation coefficient, and contextual operator weights. \\
 $W_{r,k}$ & $k$th candidate propagation operator for relation type $r$. \\
 $M_q^t$ & Temporal memory state of evidence channel $q$. \\
 $\{\lambda_q,\eta_q\}$ & Channel-specific memory decay and uncertainty growth rates. \\
 $(\hat{y}_{ij,t},\hat{u}_{ij,t})$ & Predicted trust probability and predictive uncertainty. \\
 \bottomrule
 \end{tabular}
\end{table}

Algorithm~\ref{alg:tchg} gives the computation flow of \model{}: construct evidence online, read pre-prediction memory, perform conditioned propagation, output probability and uncertainty, and write memory after prediction according to the protocol. The current candidate event is never written into memory before prediction.

\begin{algorithm}[!t]
 \caption{Tri-Trust Conditioned Heterogeneous Graph Learning (\model{})}
 \label{alg:tchg}
 \footnotesize
 \begin{algorithmic}[1]
 \Require Event stream $\mathcal{S}$, historical graph $\mathcal{G}_{<t}$, node states $\{\mathbf{h}_v\}$, component memories $\{M_q\}_{q\in\{E,B,C\}}$, candidate operators $\{W_{r,k}\}_{k=1}^{K}$
 \Ensure Trust probability $\hat{y}_{ij,t}$ and predictive uncertainty $\hat{u}_{ij,t}$
 \State Initialize training loss $\mathcal{L}_{all}\leftarrow 0$
 \For{each event $e_{ij,t}=(u_i,u_j,r,t,\mathbf{x}_{ij,t},y_{ij,t})\in\mathcal{S}$}
 \State Build historical tri-trust evidence from $\mathcal{G}_{<t}$ and encode it by Eq.~\ref{eq:evidence-encoding}
 \State Derive observation mask $o_t$
 \For{each component $q\in\{E,B,C\}$}
 \State Decay historical memory from events before $t$ by Eq.~\ref{eq:memory-decay}
 \State Enhance current evidence with pre-event memory by Eq.~\ref{eq:memory-enhanced-evidence}
 \EndFor
 \State For each incoming source $v\in\mathcal{N}_j$, compute source-side $\tau_E^{\mathrm{src}}(v,t)$ and $g_E(v\rightarrow u_j,t)$ by Eq.~\ref{eq:entity-gate}
 \State Compute pair/event-level $\rho_B(i,j,t)$ and $\boldsymbol{\alpha}_C(i,j,t)$ by Eqs.~\ref{eq:behavior-modulation} and~\ref{eq:context-operator}
 \State Propagate $\mathbf{m}_{ij}$ and update $\mathbf{h}'_j$ by Eqs.~\ref{eq:conditioned-message} and~\ref{eq:node-update}
 \State Form edge representation by Eq.~\ref{eq:edge-representation}
 \State Predict $p_{ij}$ and $u_{ij}$ by Eq.~\ref{eq:prediction-head}
 \State Set outputs $\hat{y}_{ij,t}\leftarrow p_{ij}$ and $\hat{u}_{ij,t}\leftarrow u_{ij}$
 \State Accumulate $\mathcal{L}_{all}$ by Eqs.~\ref{eq:training-objective},~\ref{eq:brier-loss}, and~\ref{eq:uncertainty-error-alignment}
 \For{each component $q\in\{E,B,C\}$}
 \State After prediction, generate candidate state by Eq.~\ref{eq:memory-candidate}
 \State If the event is allowed as future history, update memory for events after $t$ by Eq.~\ref{eq:memory-update}; otherwise keep historical memory
 \EndFor
 \EndFor
 \State Update model parameters by back-propagation on $\mathcal{L}_{all}$
 \State Optionally calibrate validation logits by Eq.~\ref{eq:affine-calibration}
 \State Return $\hat{y}_{ij,t}$ and $\hat{u}_{ij,t}$
 \end{algorithmic}
\end{algorithm}

\subsection{Tri-Trust Evidence Construction and Representation}

Dynamic user trust relationships are jointly determined by multiple sources of evidence. In order to avoid different evidence channels being mixed in a unified feature space and weakening their semantic functions, \model{} decomposes the evidence of candidate trust events into entity reliability, interaction-behavior reliability, and contextual trust, and codes them separately.

Tri-trust evidence construction serves as the input-side modeling step of TCHG. All evidence channels are computed only from information available before the target prediction time; for datasets without native trust timestamps, they are constructed only from the training graph and available user--item interaction records. The constructed channels are then used by the subsequent conditioned propagation, component memory, and uncertainty-aware prediction modules. Thus, tri-trust evidence is not treated as a standalone feature set, but as structured evidence designed for the functional mechanisms of TCHG.

For candidate events $(u_i,u_j,t)$ with native trust timestamps, the three evidence channels are computed online from the historical graph $\mathcal{G}_{<t}$. Let $\mathcal{L}_{<t}$ denote the observed user--user trust links before $t$, $\mathcal{P}_{<t}$ denote the user--item rating records $(u,a,\rho,\tau)$ before $t$, $W(t)$ denote the recent window before the target time, $\theta_l$ denote the low-rating threshold, and $\epsilon$ denote a smoothing constant. $\mathcal{P}_{u,<t}$ and $\mathcal{P}_{u,W(t)}$ denote the cumulative and recent rating records of user $u$, respectively, and $H_u$ denotes the historical trust or rating event-time set of user $u$. For Ciao and CiaoDVD, this paper does not construct pseudo trust timestamps; tri-trust evidence is instead built from the training graph and available user--item interaction records under all-users random trust-link prediction splits. Table~\ref{tab:evidence_features} gives the main feature definitions. Epinions, Ciao, and CiaoDVD do not provide explicit distrust links; therefore, randomly sampled non-connected candidate pairs are used only as training and evaluation labels, and are neither used to construct tri-trust evidence nor interpreted as explicit distrust.

All statistics read only events that have entered the historical stream before the target time. Count features are scaled by $\log(1+x)$, ratio features use additive smoothing, and time gaps use logarithmic scaling. The normalization parameters of continuous features are fitted only from training-set historical statistics and reused unchanged for validation and testing. When no history is available, the corresponding statistic is set to zero and an availability mask is appended.

\begin{table*}[!t]
 \centering
 \scriptsize
 \caption{Main feature definitions of tri-trust evidence. All features use only $\mathcal{G}_{<t}$.}
 \label{tab:evidence_features}
 \resizebox{\textwidth}{!}{%
 \begin{tabular}{@{}p{0.12\textwidth}p{0.20\textwidth}p{0.48\textwidth}p{0.13\textwidth}@{}}
 \toprule
 Evidence channel & Feature dimension & Formula or definition & Function\\
 \midrule
 \multirow{3}{*}{\makecell[c]{Entity\\reliability}} & Trust in/out degree & $d^{in}_{u,t}=|\{(v,u,\tau)\in\mathcal{L}_{<t}\}|,\quad d^{out}_{u,t}=|\{(u,v,\tau)\in\mathcal{L}_{<t}\}|$, computed for both $u_i$ and $u_j$ & Source credibility\\
 & Trusted ratio & $r^{tr}_{u,t}=(d^{in}_{u,t}+\epsilon)/(d^{in}_{u,t}+d^{out}_{u,t}+2\epsilon)$ & Long-term reputation\\
 & Activity and span & $a_{u,t}=\log(1+|\mathcal{P}_{u,<t}|+d^{in}_{u,t}+d^{out}_{u,t})$; $s_{u,t}=\log(1+\max_{\tau\in H_u}\tau-\min_{\tau\in H_u}\tau)$ & Historical sufficiency\\
 \midrule
 \multirow{4}{*}{\makecell[c]{Interaction\\behavior\\reliability}} & Co-rated item count & $n^{co}_{ij,t}=|\{a\mid (u_i,a,\rho,\tau_i),(u_j,a,\rho',\tau_j)\in\mathcal{P}_{<t}\}|$ & Behavioral overlap\\
 & Recent interaction gap & $\delta_{ij,t}=\log(1+t-\max_{\tau\in H_{ij,<t}}\tau)$, where $H_{ij,<t}$ is the time set of co-rated items or locally related interactions for the two endpoints & Recency\\
 & Low-rating ratio & $\ell_{ij,t}=(\epsilon+\sum_{(u,a,\rho,\tau)\in\mathcal{P}_{i,<t}\cup\mathcal{P}_{j,<t}}\mathbb{I}[\rho\le\theta_l])/(2\epsilon+|\mathcal{P}_{i,<t}\cup\mathcal{P}_{j,<t}|)$ & Negative behavior\\
 & Rating volatility/drift & $\sigma_{u,W(t)}=\operatorname{std}\{\rho\mid(u,a,\rho,\tau)\in\mathcal{P}_{u,W(t)}\}$; $g_{u,t}=|\bar{\rho}_{u,W(t)}-\bar{\rho}_{u,<t}|$, computed for both $u_i$ and $u_j$ & Anomalous change\\
 \midrule
 \multirow{3}{*}{\makecell[c]{Contextual\\trust}} & Relation and temporal context & Relation-type embedding $\mathbf{a}_r$; temporal position $\tilde{t}=t/T_{\max}$; previous related-event gap $\Delta t$ & Scenario condition\\
 & Local activity & $b_{ij,t}=\log(1+|\mathcal{N}_{i,W(t)}|+|\mathcal{N}_{j,W(t)}|)$ & Local environment\\
 & Low-rating trend and item context & $\Delta\ell_{ij,t}=\ell_{ij,W(t)}-\ell_{ij,<t}$; available item-category or context-type embedding, set to zero with a mask when missing & Scenario change\\
 \bottomrule
 \end{tabular}%
 }
\end{table*}

Each evidence channel in Table~\ref{tab:evidence_features} contains multiple statistical dimensions. \model{} does not manually average these dimensions. Instead, the normalized numerical features, availability masks, and available category or context embeddings in the same channel are concatenated in a fixed order into a channel-level raw feature vector:
\begin{equation}
 \label{eq:channel-feature-assembly}
 \mathbf{x}_{ij,t}^{q}
 =
 \operatorname{Concat}_{d\in\mathcal{D}_q}
 \left[
 \tilde{\mathbf{f}}_{ij,t}^{q,d},
 \boldsymbol{\mu}_{ij,t}^{q,d},
 \mathbf{c}_{ij,t}^{q,d}
 \right],
 \quad q\in\{E,B,C\},
\end{equation}
where $\mathcal{D}_q$ is the feature-dimension set of evidence channel $q$ in Table~\ref{tab:evidence_features}, $\tilde{\mathbf{f}}_{ij,t}^{q,d}$ is the scaled and training-statistics-normalized numerical feature group, $\boldsymbol{\mu}_{ij,t}^{q,d}$ is the availability mask of this dimension, and $\mathbf{c}_{ij,t}^{q,d}$ denotes an optional relation, category, or context embedding. If a dimension has no categorical embedding, this term is set to zero. For user-side statistics defined separately on $u_i$ and $u_j$, $\tilde{\mathbf{f}}_{ij,t}^{q,d}$ keeps the source value, the target value, and their difference for the directed candidate relation. For pair-level or event-level statistics, $\tilde{\mathbf{f}}_{ij,t}^{q,d}$ directly uses the corresponding normalized scalar or vector.

The three evidence channels are then aggregated by independent learnable encoders and mapped into latent representations and scalar credibility strengths:
\begin{equation}
 \label{eq:evidence-encoding}
 \begin{aligned}
 \mathbf{h}_{ij,t}^{q}
 &=
 \operatorname{MLP}_q(\mathbf{x}_{ij,t}^{q})\\
 \mathbf{z}_{ij,t}^{q}
 &=
 W_q^z\mathbf{h}_{ij,t}^{q}\\
 \tau_{ij,t}^{q}
 &=
 \sigma\!\left((\mathbf{w}_q^\tau)^\top\mathbf{h}_{ij,t}^{q}+b_q^\tau\right),
 \quad q\in\{E,B,C\}.
 \end{aligned}
\end{equation}
where $E$, $B$ and $C$ represent entity, behavior and contextual evidence respectively, $\mathbf{h}_{ij,t}^{q}$ is the intermediate representation after within-channel aggregation, $\mathbf{z}_{ij,t}^{q}$ is the corresponding latent representation, and $\tau_{ij,t}^{q}$ is the scalar credibility strength of the candidate pair or current event on evidence channel $q$. Therefore, different feature dimensions within the same evidence category are first aggregated into a channel representation $(\mathbf{z}_{ij,t}^{q},\tau_{ij,t}^{q})$; these channel representations are then used for edge-level prediction, behavior modulation, context-operator selection, and component memory. It should be distinguished that the $\tau_E^{\mathrm{src}}(v,t)$ used for message admission in Eq.~\ref{eq:entity-gate} is the source-side entity reliability strength of the incoming source user $v$, computed from source-user entity statistics in Table~\ref{tab:evidence_features}; it is not the pair-level scalar $\tau_{ij,t}^{E}$ defined here. Feature Injection Only is used as an ablation to test whether the performance gain only comes from these statistical features.

\subsection{Tri-Trust Conditioned Heterogeneous Propagation}

The core mechanism of \model{} is to embed three types of trust evidence into the heterogeneous graph propagation process, so that it is no longer just an auxiliary input feature, but controls message admission, message strength, and propagation mode respectively. In the aggregation process for target user $u_j$, entity reliability acts on each incoming source $v\rightarrow u_j$ and decides whether the source message should enter propagation. Interaction-behavior reliability and contextual trust act on the current candidate event or message edge, modulating message strength and selecting the relation propagation operator.

Entity reliability first generates a message admission gate for each incoming edge:
\begin{equation}
\label{eq:entity-gate}
g_E(v\rightarrow u_j,t) =
\sigma\left(\frac{\tau_E^{\mathrm{src}}(v,t)-\theta_E}{\beta_E}\right),
\quad v\in\mathcal{N}_j,
\end{equation}
where $\tau_E^{\mathrm{src}}(v,t)$ is the source-side entity reliability strength of incoming source user $v$ before target time $t$, computed only from the historical trust and interaction statistics of $v$; $\theta_E$ is the entity credibility threshold, and $\beta_E$ controls the smoothness of the gate. When the current candidate message is $u_i\rightarrow u_j$, $v=u_i$. Therefore, entity reliability acts as a message admission gate: messages from high-reliability sources are allowed to enter propagation with larger weights, while messages from low-reliability sources are suppressed before they are aggregated by the target node.

Interaction-behavior reliability further generates the message intensity modulation coefficient:
\begin{equation}
 \label{eq:behavior-modulation}
 \rho_B(i,j,t)
 =
 \sigma\big(\psi_B([\tau_{ij,t}^{B},\mathbf{z}_{ij,t}^{B},\mathbf{e}_{ij,t}])\big),
\end{equation}
where $\tau_{ij,t}^{B}$ and $\mathbf{z}_{ij,t}^{B}$ represent the credibility strength and latent representation of behavioral evidence for the current candidate event, respectively, $\mathbf{e}_{ij,t}$ represents edge-level auxiliary features, and $\psi_B(\cdot)$ is the behavioral modulation network. In contrast to entity reliability, which decides whether a source message should be admitted, interaction-behavior reliability continuously modulates the strength of the admitted message. A larger $\rho_B(i,j,t)$ indicates more reliable interaction behavior and preserves more propagation strength, whereas a smaller $\rho_B(i,j,t)$ weakens messages associated with unstable, anomalous, or low-confidence behavior.

Contextual trust is used to select the relation propagation operator. For each relation type $r$, \model{} maintains a relation-specific candidate operator set
\begin{equation}
 \label{eq:operator-set}
 \mathcal{W}_r=\{W_{r,1},\ldots,W_{r,K}\},
 \quad
 W_{r,k}\in\mathbb{R}^{d_h\times d_h},
\end{equation}
where $d_h$ is the hidden dimension. The $K$ candidate operators under the same relation type share the same input and output dimensions, but are independent learnable propagation matrices and can learn different propagation biases. In all experiments, $K=4$, matching the implementation setting \texttt{num\_operator\_candidates: 4}. These candidates are not manually bound to fixed semantic labels; instead, the supervised prediction loss and the operator-diversity regularization term $\mathcal{L}_{div}$ in Eq.~\ref{eq:training-objective}, implemented by \texttt{operator\_diversity}, encourage them to differentiate automatically. For interpretation, their learned functions can be summarized as basic relation propagation, stable-context reinforcement, temporal or phase-drift adaptation, and conservative propagation under noisy or low-confidence contexts.

For a concrete event $e=(u_i,u_j,r,t)$, the model does not hard select a single candidate operator. Instead, it embeds the context representation, context credibility strength and relation type into the operator selection network and obtains a soft distribution over the candidate operators:
\begin{equation}
 \label{eq:context-operator}
 \begin{array}{@{}l@{}}
 \boldsymbol{\alpha}_C
 =
 \operatorname{softmax}
 \left(
 \psi_C([\mathbf{z}_C,\tau_C,\mathbf{a}_r])
 \right),\\[2pt]
 \displaystyle\sum_{k=1}^{K}\alpha_{C,k}
 =
 1.
 \end{array}
\end{equation}
where $\alpha_{C,k}$ represents the selection weight of the $k$th candidate propagation operator, and $\mathbf{a}_r$ represents the embedding of relation type $r$. The context-conditioned propagation operator for the current event is then the weighted mixture
\begin{equation}
 \label{eq:context-mixed-operator}
 W^{C}_{r,ij,t}
 =
 \sum_{k=1}^{K}
 \alpha_{C,k}W_{r,k}.
\end{equation}
Thus the same relation type can use different propagation transformations under different timestamps, item categories, user activity levels, and contextual reliability conditions. The final conditional message from the source user to the target user is defined as:
\begin{equation}
 \label{eq:conditioned-message}
 \mathbf{m}_{ij}
 =
 g_E(u_i\rightarrow u_j,t)\rho_B(i,j,t)
 (\mathbf{h}_i W^{C}_{r,ij,t}).
\end{equation}

The operator selector is a learnable soft-assignment mechanism. When history is sufficient, context is consistent, or credibility is high, the weights may concentrate on operators that strengthen semantic propagation. When history is sparse, context is missing, or noise is strong, the weights remain smoother and lean toward conservative propagation. The relation embedding $\mathbf{a}_r$ enters the selector, allowing the same contextual features to trigger different operator mixtures for different edge types.

If context-operator selection is disabled, $\boldsymbol{\alpha}_C$ is replaced by the uniform distribution $\alpha_{C,k}=1/K$, and $W^{C}_{r,ij,t}$ degenerates into the average mixture of candidate operators. This setting keeps relation-aware propagation but removes context-driven propagation-mode switching, corresponding to the w/o Context Operator Selection ablation.

All messages directed to target user $u_j$ are aggregated at the target node, and a new node representation is obtained through residual update:
\begin{equation}
 \label{eq:node-update}
 \begin{aligned}
 \bar{\mathbf{m}}_j
 &=
 \frac{1}{|\mathcal{N}_j|}
 \sum_{i\in\mathcal{N}_j}\mathbf{m}_{ij}\\
 \mathbf{h}'_j
 &=
 \mathrm{LN}\left(
 \mathbf{h}_j+
 \psi_h([\mathbf{h}_j,\bar{\mathbf{m}}_j])
 \right),
 \end{aligned}
\end{equation}
where $\mathcal{N}_j$ is the neighbor set pointing to the target user $u_j$, $\psi_h(\cdot)$ is the node update network, and $\mathrm{LN}(\cdot)$ denotes layer normalization. In implementation, the summation in Eq.~\ref{eq:node-update} is performed by an \texttt{index\_add\_}-based mean aggregation over destination node indices and normalized by destination-specific message counts.

\subsection{Component-Decoupled Temporal Trust Memory}

Dynamic trust prediction not only depends on the tri-trust evidence of the current candidate event, but is also continuously affected by the historical trust state. Since entity reliability, interaction-behavior reliability, and contextual trust have different temporal stability, \model{} maintains three temporal memory states $M_E$, $M_B$, and $M_C$. In general, entity reliability has stronger historical accumulation, behavior reliability is affected by both long-term interaction patterns and recent behavioral changes, and contextual trust fluctuates more rapidly with time, relational environments, and local scenes. Therefore, component-decoupled memory is not simply an additional temporal branch; it prevents volatile contextual evidence from overwriting long-term entity reliability and prevents slow entity states from masking recent behavioral anomalies. Under this design, $M_C$ is not expected to become a high-norm long-term memory. Its role is to preserve necessary contextual history residuals, while the main effect of context on current propagation occurs through operator selection in Eq.~\ref{eq:context-operator}.

For any evidence component $q\in\{E,B,C\}$, \model{} first performs non-uniform decay on the historical memory according to the time interval $\Delta t$, and estimates the uncertainty of the memory state:
\begin{equation}
 \label{eq:memory-decay}
 \begin{aligned}
 \bar{M}_{q}^{t}
 &=
 \exp(-\lambda_q\Delta t)M_{q}^{t-1}\\
 U_{q}^{t}
 &=
 1-\exp(-\eta_q\Delta t),
 \end{aligned}
\end{equation}
where $\lambda_q$ represents the temporal decay rate of memory of type $q$, and $\eta_q$ represents the corresponding uncertainty growth rate. Different components use different $\lambda_q$ and $\eta_q$ to characterize the differences in history retention and uncertainty accumulation among the three types of trust evidence.

When predicting the current event, the model only uses the decayed historical memory $\bar{M}_q^t$, which is obtained from events before the target time. After the prediction of the current event is completed, if the event is allowed to be used as part of the subsequent history stream, the model then uses GRU to generate candidate memory states, and merges them with the attenuated historical memory through the update gate; when the current event is not observed or is not allowed to be written into the history, the memory only performs temporal decay. The post-prediction writing process can be expressed as:
\begin{equation}
 \label{eq:memory-update}
 M_q^t
 =
 (1-o_t)\bar{M}_q^t
 +
 o_t
 \left[
 \gamma_q^t\tilde{M}_q^t
 +(1-\gamma_q^t)\bar{M}_q^t
 \right],
\end{equation}
where $o_t$ is the observation mask, $\tilde{M}_q^t$ is the candidate memory, and $\gamma_q^t$ is the update gate. We set $o_t=1$ only when the current event is allowed by the protocol to enter future history. Observed training events can be written after prediction, whereas validation and test candidate events default to $o_t=0$ and their labels are not written into memory. Candidate memory is generated by both the current evidence representation and edge-level features:
\begin{equation}
 \label{eq:memory-candidate}
 \begin{aligned}
 \tilde{M}_q^t
 &=
 \mathrm{GRU}_q(\mathbf{r}_q^t,\bar{M}_q^t)\\
 \mathbf{r}_q^t
 &=
 P_q([\mathbf{z}_{ij,t}^{q},\mathbf{e}_{ij,t}]).
 \end{aligned}
\end{equation}
where $P_q(\cdot)$ is the component-specific projection function. Eqs.~\ref{eq:memory-update} and~\ref{eq:memory-candidate} describe post-prediction state writing; validation and test labels are never used for memory updates before prediction. This mechanism avoids indiscriminate event writing.

The decayed history memory used for the current prediction is further fed back into the current evidence representation:
\begin{equation}
 \label{eq:memory-enhanced-evidence}
 \widehat{\mathbf{z}}_{ij,t}^{q}
 =
 \mathrm{LN}\left(
 \mathbf{z}_{ij,t}^{q}
 +(1-U_q^t)A_q\bar{M}_q^t
 \right),
 \quad
 q\in\{E,B,C\},
\end{equation}
where $A_q$ is the component-specific linear mapping matrix, and $\widehat{\mathbf{z}}_{ij,t}^{q}$ is the evidence representation after memory enhancement. This formula shows that low-uncertainty historical memories can strengthen current evidence representation, while the impact of high-uncertainty memories will be relatively weakened. As a result, the component-decoupled memory module enables entities, behaviors, and contextual evidence to evolve on their respective time scales, thereby enhancing the model's ability to depict long-term historical states and recent dynamic changes.

\subsection{Uncertainty-Aware Prediction and Calibration}

After completing the tri-trust conditioned propagation and evidence enhancement based on historical memory, \model{} combines the source user representation, the target user representation, the memory-enhanced tri-trust evidence representation, pre-prediction attenuated historical memory, memory uncertainty and edge-level features into an edge-level prediction representation:
\begin{equation}
 \label{eq:edge-representation}
 \mathbf{r}_{ij,t}
 =
 [
 \mathbf{h}'_i,
 \mathbf{h}'_j,
 \widehat{\mathbf{z}}_{ij,t}^{E},
 \widehat{\mathbf{z}}_{ij,t}^{B},
 \widehat{\mathbf{z}}_{ij,t}^{C},
 \bar{M}_E^t,\bar{M}_B^t,\bar{M}_C^t,
 U_E^t,U_B^t,U_C^t,
 \mathbf{e}_{ij,t}
 ].
\end{equation}
The terms in Eq.~\ref{eq:edge-representation} play different roles: $\mathbf{h}'_i$ and $\mathbf{h}'_j$ encode propagated structural states, while $\widehat{\mathbf{z}}_{ij,t}^{E/B/C}$ retain current tri-trust evidence, $\bar{M}_{E/B/C}^{t}$ provide pre-prediction historical states, $U_{E/B/C}^{t}$ encode memory reliability, and $\mathbf{e}_{ij,t}$ preserves auxiliary edge information not covered by the tri-trust channels. Multi-dimensional statistics are first organized into channel representations and then enhanced through propagation and memory, rather than being fed as scattered raw features directly into the classifier.
The prediction head outputs the trust logit $s_{ij}$ and the uncertainty logit $a_{ij}$ based on $\mathbf{r}_{ij,t}$, and obtains the trust probability and uncertainty scores through the sigmoid function:
\begin{equation}
 \label{eq:prediction-head}
 \begin{aligned}
 p_{ij}
 &=
 \sigma(s_{ij})\\
 u_{ij}
 &=
 \sigma(a_{ij}).
 \end{aligned}
\end{equation}
where $p_{ij}$ is the predicted probability that user $u_i$ trusts user $u_j$, and $u_{ij}$ is the model uncertainty estimate. The uncertainty score does not directly modify the probability; it represents sample-level risk.

The training objective of \model{} includes binary prediction, probability quality, uncertainty regularization, error-uncertainty alignment, and operator diversity:
\begin{equation}
 \label{eq:training-objective}
 \mathcal{L}
 =
 \mathcal{L}_{BCE}
 +\lambda_{brier}\mathcal{L}_{brier}
 +\lambda_{unc}\mathcal{L}_{unc}
 +\lambda_{uerr}\mathcal{L}_{uerr}
 +\lambda_{div}\mathcal{L}_{div}.
\end{equation}
where $\mathcal{L}_{BCE}$ is binary cross-entropy; $\mathcal{L}_{brier}$ improves probabilistic quality; $\mathcal{L}_{unc}$ and $\mathcal{L}_{uerr}$ constrain sample-level uncertainty and its consistency with actual error; $\mathcal{L}_{div}$ prevents context-operator selection from degenerating.

Brier loss is defined as~\cite{brier1950}:
\begin{equation}
 \label{eq:brier-loss}
 \mathcal{L}_{brier}
 =
 \frac{1}{|\mathcal{B}|}
 \sum_{(i,j)\in\mathcal{B}}
 (p_{ij}-y_{ij})^2,
\end{equation}
where $\mathcal{B}$ is the training batch and $y_{ij}\in\{0,1\}$ is the true label. Error-uncertainty alignment is defined as:
\begin{equation}
 \label{eq:uncertainty-error-alignment}
 \mathcal{L}_{uerr}
 =
 \frac{1}{|\mathcal{B}|}
 \sum_{(i,j)\in\mathcal{B}}
 \left(
 u_{ij}-|p_{ij}-y_{ij}|
 \right)^2,
\end{equation}
This term encourages predictive uncertainty to match actual prediction error. Posterior calibration learns a monotonic affine logit calibration on the validation set:
\begin{equation}
 \label{eq:affine-calibration}
 p_{ij}^{cal}
 =
 \sigma(\alpha_{cal}s_{ij}+\beta_{cal}),
\end{equation}
where $\alpha_{cal}$ and $\beta_{cal}$ are validation-set calibration parameters and are fixed at test time; $\alpha_{cal}>0$ preserves ranking monotonicity. MRR, AP, and AUC are computed from pre-calibration outputs, while $p_{ij}^{cal}$ is used only for reliability metrics such as ECE, Brier Score, and NLL.

\section{Experiments}

This section first describes the experimental protocol, data splits, baselines, and metrics, and then reports overall performance, ablations, temporal-memory analysis, efficiency, and robustness results.

\subsection{Experimental Protocol and Fairness Audit}

\subsubsection{Datasets and Data Preparation}

To evaluate the effectiveness of \model{} for user trust prediction, this paper selects three public social trust datasets: Epinions, Ciao, and CiaoDVD. All three datasets contain user--item rating records and user--user trust relations, which support user interaction modeling, contextual evidence construction, and trust link prediction. Table~\ref{tab:dataset} gives the basic statistics of the datasets and experimental protocols.

The negative sampling protocol follows common practice in trust prediction. Observed user--user trust links are positive samples, and non-connected user pairs are randomly sampled as negatives at a 1:1 ratio, consistent with CAT~\cite{wang2026cat} and KGTrust~\cite{yu2023}. After negative sampling, Epinions, Ciao, and CiaoDVD contain 622316, 115088, and 80266 candidate samples, respectively. The same protocol is used for \model{} and all baselines, ensuring identical candidate-edge distribution, label definition, and positive-negative ratio.

In terms of time attributes, Epinions contains both rating timestamps and trust timestamps, and is therefore used as the full-timestamp dynamic trust prediction setting. For Epinions, we further distinguish observed-user and unobserved-user test cases. Ciao and CiaoDVD contain rating timestamps but do not provide native trust timestamps. To avoid protocol assumptions caused by artificial trust-time construction, this paper does not generate pseudo trust timestamps for Ciao or CiaoDVD. Instead, candidate trust links from all users are randomly split into training, validation, and test sets under the 70\%-15\%-15\% and 80\%-10\%-10\% ratios. Thus, Epinions supports the strict dynamic trust modeling claim, while Ciao and CiaoDVD provide additional all-users random trust-link prediction splits without native trust timestamps. For Ciao and CiaoDVD, tri-trust evidence is constructed only from the training graph and available user--item interaction records; validation and test trust labels are not used for feature construction or memory writing. The dataset statistics and protocol audit results are shown in Table~\ref{tab:dataset}.

\begin{table*}[!t]
 \centering
 \scriptsize
 \caption{Dataset and experimental protocol statistics.}
 \label{tab:dataset}
 \resizebox{\textwidth}{!}{%
 \begin{tabular}{@{}lcccccccccc@{}}
 \toprule
 Dataset& Users& Items& Ratings& Trust links& Context types& Negative sampling ratio& Candidate samples& Rating timestamp& Trust timestamp& Split protocol\\
 \midrule
 Epinions & 9163 & 12573 & 265189 & 311158 & 25 & 1.0 & 622316 & yes& yes& temporal\\
 Ciao & 2378 & 16861 & 36065 & 57544 & 6 & 1.0 & 115088 & yes& no& all-users random trust-link prediction\\
 CiaoDVD & 19533 & 16121 & 72665 & 40133 & 17 & 1.0 & 80266 & yes& no& all-users random trust-link prediction\\
 \bottomrule
 \end{tabular}%
 }
\end{table*}

The protocol audit shows that all datasets use unified candidate-edge construction, balanced negative sampling, evaluation metrics, validation-only threshold selection, and posterior calibration. The split strategy is selected according to trust-time availability: Epinions uses temporal splits, whereas Ciao and CiaoDVD use all-users random trust-link prediction splits. This design avoids introducing pseudo trust timestamps on Ciao and CiaoDVD while reducing protocol inconsistency and future-information leakage.

\subsubsection{Baseline Models}

We compare Linear, Guardian, TrustGNN, CAT, HAN, RGCN, and HGT. Linear tests the basic discriminative power of candidate-edge features. Guardian, TrustGNN, and CAT are trust-prediction models, with CAT being closest to our task. HAN, RGCN, and HGT represent general graph neural networks and heterogeneous graph models~\cite{lin2020,huo2024,wang2026cat,wang2019han,schlichtkrull2018,hu2020}.

All baselines are adapted under unified candidate edges, positive/negative samples, task mapping, and metrics. Epinions uses the same temporal splits, and Ciao/CiaoDVD use the same all-users random trust-link prediction splits. Tri-trust evidence channels are used only by TCHG because they are coupled with its message admission, behavior modulation, context-conditioned operator selection, component memory, and uncertainty-aware prediction modules. Supplying these channels to all baselines would change their original input assumptions and create feature-augmented variants of those models. To examine whether the performance gain comes merely from these constructed evidence features, we include Feature Injection Only and Attention Only ablations under the same candidate-edge protocol, data splits, and evaluation metrics.

\subsubsection{Evaluation Metrics}

The main metrics reported in the overall-performance tables are MRR, AP, and AUC. MRR measures the rank of true trust edges in the candidate set, AP measures positive-edge retrieval quality, and AUC measures positive-negative discrimination. F1 and Accuracy are reported only as threshold-based classification supplements in the targeted temporal-memory analysis in Table~\ref{tab:medium_gap}; they are not included in the overall-performance tables. MRR, AP, and AUC use pre-calibration outputs; posterior calibrated probabilities are used only for reliability metrics such as ECE, Brier Score, and NLL.

\subsubsection{Parameter Settings and Implementation Details}

All models in this paper are trained and tested based on the same preprocessed event files, candidate edge sets, and evaluation protocols. The experiment adopts two data partitioning protocols of 70\%-15\%-15\% and 80\%-10\%-10\%. All models use the same training set, validation set and test set partitioning. Model selection, classification threshold determination, and posterior calibration parameter learning are all completed on the validation set without using any test set information.

The hyperparameters in this paper are all determined by the validation set. We set the maximum number of epochs to 20, and stop early if the AUC of the validation set does not improve for 6 consecutive epochs; the optimizer is AdamW, the learning rate is 0.001, the weight decay is 0.0001, the dropout is 0.15, the hidden dimension is 64, the trust representation dimension and the memory dimension are both 24, the number of propagation layers is 2, and the number of candidate heterogeneous operators is 4. The unique hyperparameter settings in this paper are as follows: the loss weights are set to $\lambda_{\mathrm{BCE}}=1.0$, $\lambda_{\mathrm{Brier}}=0.20$, $\lambda_{\mathrm{unc}}=0.20$, $\lambda_{\mathrm{uerr}}=0.10$, $\lambda_{\mathrm{div}}=0.002$; the entity-gate parameters are set to $\theta_E=0.45$ and $\beta_E=0.20$; the temporal decay coefficients of the three memory components of E/B/C are set to $\lambda_E=0.015$, $\lambda_B=0.045$, $\lambda_C=0.090$, the uncertainty growth coefficient is set to $\eta_E=0.025$, $\eta_B=0.060$, $\eta_C=0.120$. The loss weights are not manually tuned on the test set. With $\lambda_{\mathrm{BCE}}=1.0$ fixed as the main ranking loss weight, they are selected by a coarse validation-grid search: $\lambda_{\mathrm{Brier}}$ and $\lambda_{\mathrm{unc}}$ are chosen from $\{0.05,0.10,0.20,0.50\}$, $\lambda_{\mathrm{uerr}}$ from $\{0.05,0.10,0.20\}$, and $\lambda_{\mathrm{div}}$ from $\{0,0.001,0.002,0.005\}$. The final combination is selected when validation ranking metrics do not decrease and ECE, Brier Score, and NLL improve.

To evaluate generalization under different historical availability, this paper divides candidate user pairs into observed-user and unobserved-user cases only for the Epinions test set. If both endpoints of a candidate pair appear in the training graph and have available historical trust or interaction evidence before the target prediction time, the sample is treated as observed-user. If at least one endpoint does not appear in the training graph, or at least one endpoint lacks available historical evidence before the target prediction time, the sample is treated as unobserved-user. This split is used only for Epinions test evaluation and does not change the training data, negative sampling strategy, or model input protocol. Since Ciao and CiaoDVD lack native trust timestamps, they are evaluated under all-users random trust-link prediction splits and report overall results over all test candidate edges, without observed-user or unobserved-user subdivision.

In order to reduce the impact of random initialization, negative sampling, and randomness in the training process on the results, each set of experiments was independently repeated for five rounds under the same protocol, and the tables report the average of the five rounds of experimental results.

The experimental hardware is NVIDIA RTX 3060 Laptop GPU. The efficiency experiment counts the average single-epoch training time, amount of trainable parameters, and peak GPU memory of different models in the same hardware environment.

\subsection{Overall Performance and Generalization Ability}

Tables~\ref{tab:overall_epinions}--\ref{tab:overall_ciaodvd} compare \model{} with all baselines. On Epinions, \model{} reaches 0.8811 MRR, 0.9832 AP, and 0.9866 AUC in the 80\%-10\%-10\% observed-user setting, and 0.7018 MRR, 0.9755 AP, and 0.9690 AUC in the unobserved-user setting, clearly outperforming CAT. These results show that, when complete trust timestamps are available, tri-trust conditioned propagation and component memory improve ranking quality for both history-rich and history-sparse samples.

The advantage remains on Ciao and CiaoDVD under all-users random trust-link prediction splits. On Ciao, \model{} obtains 0.6434 and 0.6499 MRR under the 70\%-15\%-15\% and 80\%-10\%-10\% splits, respectively, improving over CAT by about 53.3\% and 48.8\%. On CiaoDVD, \model{} reaches 0.9262 and 0.9325 MRR, higher than CAT's 0.8418 and 0.8608. These results show that tri-trust evidence does not rely on artificial trust timestamps and provides additional all-users trust-link prediction evidence when native trust timestamps are unavailable.

Epinions provides direct evidence for dynamic trust prediction with native trust timestamps, while Ciao and CiaoDVD provide additional all-users trust-link prediction evidence when native trust timestamps are unavailable.

\begin{table}[!htbp]
 \centering
 \scriptsize
 \setlength{\tabcolsep}{3pt}
 \caption{Overall performance on the Epinions dataset.}
 \label{tab:overall_epinions}
 \resizebox{\columnwidth}{!}{%
 \begin{tabular}{@{}llcccccc@{}}
 \toprule
 Task& Model& \multicolumn{3}{c}{70\%-15\%-15\%} & \multicolumn{3}{c}{80\%-10\%-10\%}\\
 \cmidrule(lr){3-5}\cmidrule(l){6-8}
 & & MRR & AP & AUC & MRR & AP & AUC\\
 \midrule
 Observed-user& Linear & 0.3777 & 0.8217 & 0.8879 & 0.3549 & 0.8274 & 0.8901\\
 & Guardian & 0.2970 & 0.8252 & 0.9132 & 0.4825 & 0.9080 & 0.9403\\
 & HAN & 0.4047 & 0.8557 & 0.9299 & 0.4103 & 0.8717 & 0.9327\\
 & RGCN & 0.4057 & 0.8482 & 0.9165 & 0.4765 & 0.8747 & 0.9169\\
 & TrustGNN & 0.4563 & 0.8522 & 0.9108 & 0.4969 & 0.8836 & 0.9267\\
 & HGT & 0.4947 & 0.9181 & 0.9425 & 0.6028 & 0.9472 & 0.9603\\
 & CAT & 0.6028 & 0.9379 & 0.9674 & 0.6758 & 0.9599 & 0.9771\\
 & TCHG & 0.7171 & 0.9735 & 0.9782 & 0.8811 & 0.9832 & 0.9866\\
 \midrule
 Unobserved-user& Linear & 0.2420 & 0.8977 & 0.7845 & 0.2555 & 0.9286 & 0.8086\\
 & Guardian & 0.1398 & 0.7923 & 0.5946 & 0.1532 & 0.8461 & 0.6209\\
 & HAN & 0.2536 & 0.9234 & 0.8535 & 0.1932 & 0.9303 & 0.8413\\
 & RGCN & 0.2794 & 0.8078 & 0.7973 & 0.1674 & 0.8482 & 0.6101\\
 & TrustGNN & 0.2379 & 0.8796 & 0.7108 & 0.2436 & 0.9181 & 0.7628\\
 & HGT & 0.2620 & 0.9300 & 0.8588 & 0.2843 & 0.9422 & 0.8570\\
 & CAT & 0.4049 & 0.9321 & 0.8920 & 0.4282 & 0.9594 & 0.8800\\
 & TCHG & 0.4767 & 0.9618 & 0.9678 & 0.7018 & 0.9755 & 0.9690\\
 \bottomrule
 \end{tabular}%
 }
\end{table}

\begin{table}[!htbp]
 \centering
 \scriptsize
 \setlength{\tabcolsep}{3pt}
 \caption{Overall performance on Ciao under all-users random trust-link prediction splits.}
 \label{tab:overall_ciao}
 \resizebox{\columnwidth}{!}{%
 \begin{tabular}{@{}lcccccc@{}}
 \toprule
 Model& \multicolumn{3}{c}{70\%-15\%-15\%} & \multicolumn{3}{c}{80\%-10\%-10\%}\\
 \cmidrule(lr){2-4}\cmidrule(l){5-7}
 & MRR & AP & AUC & MRR & AP & AUC\\
 \midrule
 Linear & 0.2410 & 0.7916 & 0.7891 & 0.2416 & 0.8007 & 0.8142\\
 Guardian & 0.2375 & 0.8247 & 0.8586 & 0.2423 & 0.8161 & 0.8579\\
 HAN & 0.2579 & 0.8373 & 0.8482 & 0.2641 & 0.8401 & 0.8535\\
 RGCN & 0.2120 & 0.7962 & 0.8221 & 0.2737 & 0.8229 & 0.8099\\
 TrustGNN & 0.2001 & 0.7873 & 0.8014 & 0.2035 & 0.7952 & 0.8189\\
 HGT & 0.2891 & 0.8407 & 0.8575 & 0.3137 & 0.8511 & 0.8672\\
 CAT & 0.4197 & 0.9274 & 0.9415 & 0.4367 & 0.9371 & 0.9220\\
 TCHG & 0.6434 & 0.9613 & 0.9564 & 0.6499 & 0.9632 & 0.9597\\
 \bottomrule
 \end{tabular}%
 }
\end{table}

\begin{figure}[!htbp]
 \centering
 \includegraphics[width=\columnwidth]{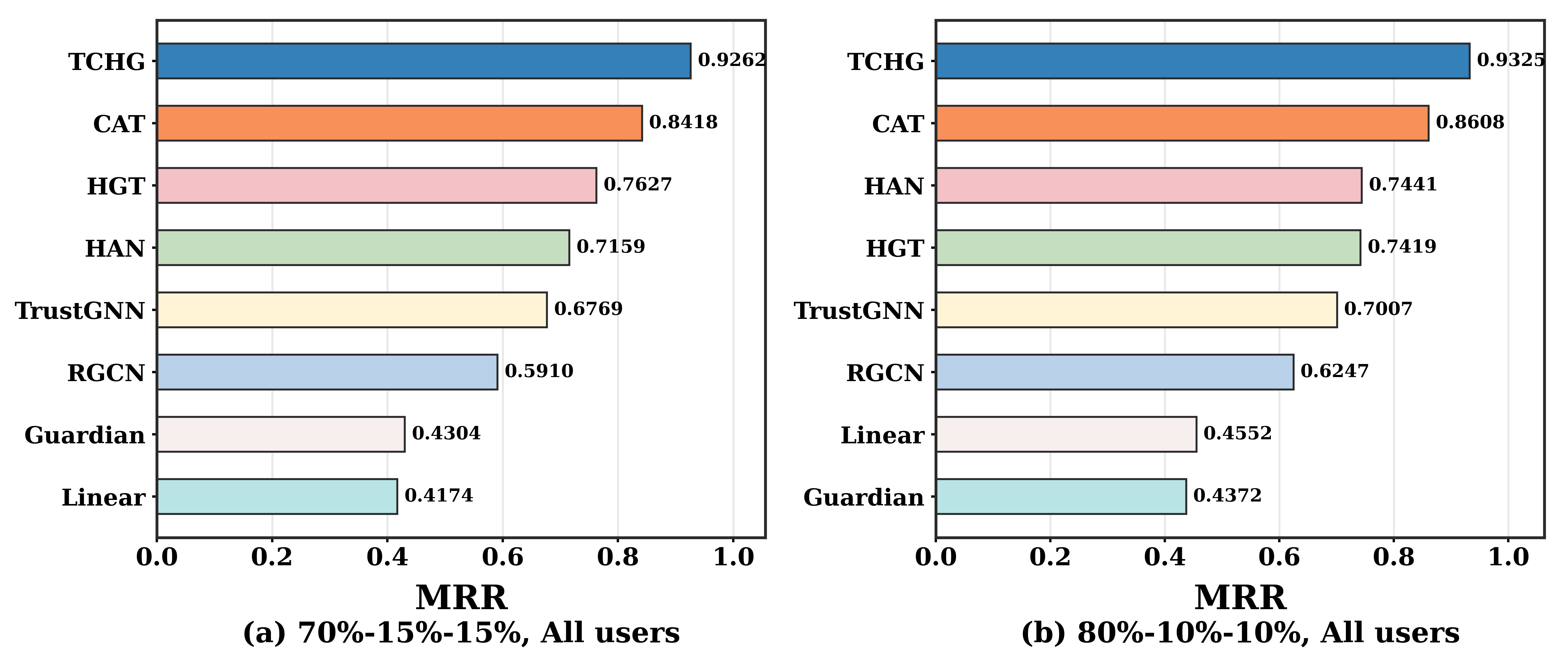}
 \caption{MRR comparison on CiaoDVD under all-users random trust-link prediction splits. Panels (a) and (b) denote the 70\%-15\%-15\% and 80\%-10\%-10\% splits, respectively.}
 \label{fig:mrr_ciaodvd}
\end{figure}

\begin{table}[!htbp]
 \centering
 \scriptsize
 \setlength{\tabcolsep}{3pt}
 \caption{Overall performance on CiaoDVD under all-users random trust-link prediction splits.}
 \label{tab:overall_ciaodvd}
 \resizebox{\columnwidth}{!}{%
 \begin{tabular}{@{}lcccccc@{}}
 \toprule
 Model& \multicolumn{3}{c}{70\%-15\%-15\%} & \multicolumn{3}{c}{80\%-10\%-10\%}\\
 \cmidrule(lr){2-4}\cmidrule(l){5-7}
 & MRR & AP & AUC & MRR & AP & AUC\\
 \midrule
 Linear & 0.4174 & 0.9349 & 0.9535 & 0.4552 & 0.9373 & 0.9545\\
 Guardian & 0.4304 & 0.9102 & 0.9289 & 0.4372 & 0.9146 & 0.9331\\
 HAN & 0.7159 & 0.9653 & 0.9626 & 0.7441 & 0.9465 & 0.9654\\
 RGCN & 0.5910 & 0.9459 & 0.7986 & 0.6247 & 0.9562 & 0.8112\\
 TrustGNN & 0.6769 & 0.9481 & 0.9588 & 0.7007 & 0.9588 & 0.9690\\
 HGT & 0.7627 & 0.9655 & 0.9655 & 0.7419 & 0.9764 & 0.9613\\
 CAT & 0.8418 & 0.9764 & 0.9775 & 0.8608 & 0.9820 & 0.9823\\
 TCHG & 0.9262 & 0.9889 & 0.9883 & 0.9325 & 0.9891 & 0.9887\\
 \bottomrule
 \end{tabular}%
 }
\end{table}
Figures~\ref{fig:mrr_epinions}--\ref{fig:mrr_ciaodvd} show MRR results by dataset. The Epinions figure contains observed-user and unobserved-user results under the two temporal splits; the Ciao and CiaoDVD figures show all-users random trust-link prediction splits under the 70\%-15\%-15\% and 80\%-10\%-10\% settings.

\par\noindent
\begin{minipage}{\columnwidth}
 \centering
 \includegraphics[width=\columnwidth]{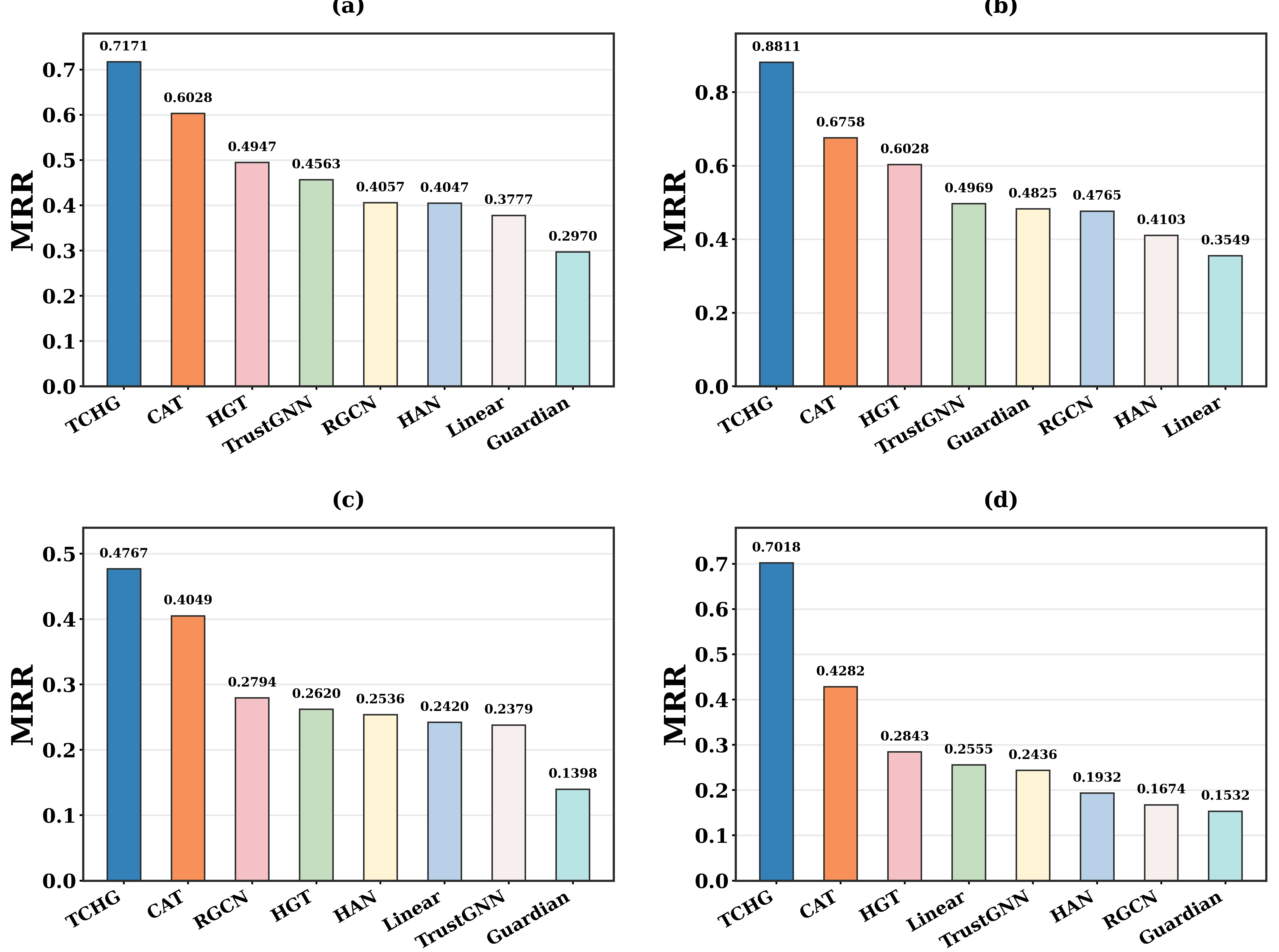}
 \captionof{figure}{MRR comparison on the Epinions dataset. Subfigures (a) -- (d) respectively represent observed-user results under the 70\%-15\%-15\% split, observed-user results under the 80\%-10\%-10\% split, unobserved-user results under the 70\%-15\%-15\% split, and unobserved-user results under the 80\%-10\%-10\% split.}
 \label{fig:mrr_epinions}
\end{minipage}
\par

\subsection{Ablation Experiments on the Epinions Dataset}

We conduct ablation experiments on Epinions under the 70\%-15\%-15\% split. All variants use the same candidate edges, data split, and metrics; the Full \model{} ranking results match the corresponding Epinions results in Table~\ref{tab:overall_epinions}.

\begin{figure}[!htbp]
 \centering
 \includegraphics[width=\columnwidth]{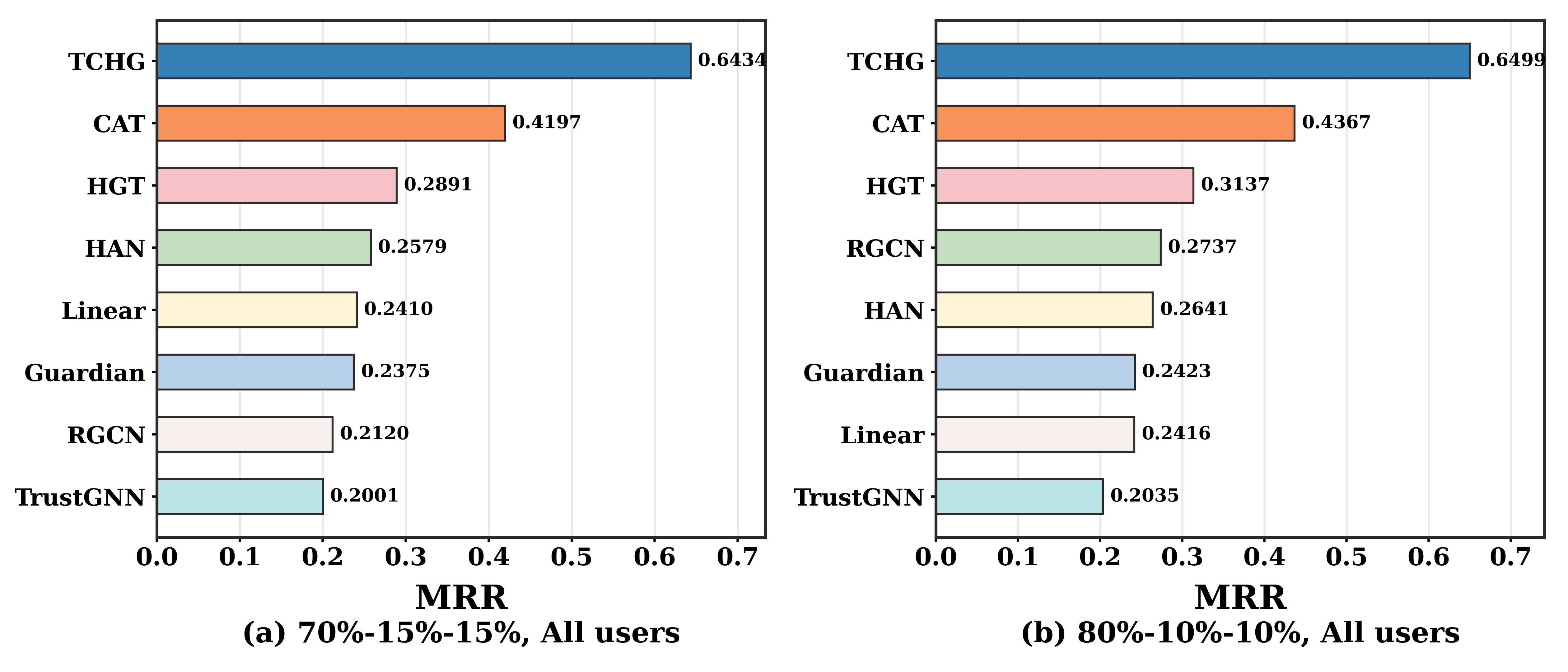}
 \caption{MRR comparison on Ciao under all-users random trust-link prediction splits. Panels (a) and (b) denote the 70\%-15\%-15\% and 80\%-10\%-10\% splits, respectively.}
 \label{fig:mrr_ciao}
\end{figure}

\subsubsection{Impact of Tri-Trust Conditioned Propagation}

Table~\ref{tab:ablation_propagation} tests whether tri-trust evidence should act as propagation control variables. Attention Only uses ordinary attention over neighbors, Feature Injection Only only concatenates tri-trust features, and the other variants remove entity admission, behavior modulation, or context-operator selection.

This ablation separates the effect of tri-trust evidence construction from the effect of evidence-controlled propagation.

Feature Injection Only and Attention Only both produce lower MRR than Full \model{}, showing that the MRR gains come from propagation control rather than feature stacking or generic attention. In the unobserved-user setting, Full \model{} improves MRR from 0.4027 to 0.4767 over Feature Injection Only, indicating that functional propagation better alleviates limited historical evidence. Removing entity admission, behavior modulation, or context-operator selection also lowers MRR, confirming that the three evidence types provide complementary controls for top-ranked trust-link retrieval.

\begin{table*}[!t]
 \centering
 \scriptsize
 \caption{Tri-trust conditioned propagation ablation experiments on the Epinions dataset.}
 \label{tab:ablation_propagation}
 \resizebox{\textwidth}{!}{%
 \begin{tabular}{@{}lccccccc@{}}
 \toprule
 Model& Observed MRR& Observed AP& Observed AUC& Unobserved MRR& Unobserved AP& Unobserved AUC& Effect\\
 \midrule
 Full \model{} & 0.7171 & 0.9735 & 0.9782 & 0.4767 & 0.9618 & 0.9678 & Complete model\\
 Attention Only & 0.4145 & 0.8810 & 0.9446 & 0.2773 & 0.9501 & 0.9167 & General attention control\\
 Feature Injection Only & 0.5610 & 0.9326 & 0.9679 & 0.4027 & 0.9516 & 0.9486 & Feature splicing control\\
 w/o Entity Admission Gate & 0.5579 & 0.9478 & 0.9602 & 0.4053 & 0.9550 & 0.9540 & Remove message admission\\
 w/o Behavior Reliability Modulation & 0.5641 & 0.9494 & 0.9618 & 0.4208 & 0.9418 & 0.9567 & Remove reliability modulation\\
 w/o Context Operator Selection & 0.6095 & 0.9559 & 0.9605 & 0.4108 & 0.9421 & 0.9606 & Remove operator selection\\
 \bottomrule
 \end{tabular}%
 }
\end{table*}

\subsubsection{Impact of Component-Decoupled Temporal Trust Memory}

Table~\ref{tab:ablation_memory} evaluates temporal-memory variants. w/o Masked Memory removes the memory branch; w/o Component-wise Memory removes independent entity, behavior, and context memories; Uniform Decay Memory uses the same temporal decay for all evidence types; w/o Reconnect Correction removes historical-state correction.

Removing memory, sharing component memory, using uniform decay, or removing historical-state correction all reduce observed-user and unobserved-user MRR. w/o Component-wise Memory causes the largest MRR degradation, showing that independent memory states are critical for top-ranked trust-link retrieval. Uniform Decay Memory increases ECE and Brier Score, indicating that unified decay harms probability quality. Component memories, non-uniform decay, and historical-state correction jointly preserve separable dynamic trust states.

\begin{table*}[!t]
 \centering
 \scriptsize
 \caption{Temporal trust memory ablation experiments on the Epinions dataset.}
 \label{tab:ablation_memory}
 \resizebox{\textwidth}{!}{%
 \begin{tabular}{@{}lccccccc@{}}
 \toprule
 Model& Observed MRR& Observed AUC& Unobserved MRR& Unobserved AUC& ECE $\downarrow$ & Brier $\downarrow$ & Effect\\
 \midrule
 Full \model{} & 0.7171 & 0.9782 & 0.4767 & 0.9678 & 0.0158 & 0.0272 & Complete temporal memory\\
 w/o Masked Memory & 0.6101 & 0.9716 & 0.4051 & 0.9536 & 0.0405 & 0.0394 & Remove memory branch\\
 w/o Component-wise Memory & 0.5239 & 0.9649 & 0.3727 & 0.9382 & 0.0941 & 0.1450 & Remove component-decoupled memory\\
 Uniform Decay Memory & 0.5989 & 0.9704 & 0.4171 & 0.9594 & 0.0540 & 0.0417 & Use uniform temporal decay\\
 w/o Reconnect Correction & 0.5664 & 0.9699 & 0.3810 & 0.9408 & 0.0483 & 0.0474 & Remove historical status correction\\
 \bottomrule
 \end{tabular}%
 }
\end{table*}

\subsubsection{Influence of Uncertainty-Aware Prediction and Calibration}

Table~\ref{tab:ablation_calibration} compares Full \model{}, w/o Uncertainty Calibration, and w/o Post-hoc Calibration. The table reports only ECE, Brier Score, and NLL; main ranking metrics use pre-calibration outputs and are not repeated here.

ECE is mainly improved by validation-set posterior logit calibration. Uncertainty constraints contribute less to ECE, but removing them increases Brier Score and NLL, showing that they improve probability error and negative log-likelihood. Affine logit calibration is monotonic, so it changes probability scale rather than candidate-edge ordering.

\begin{table}[!htbp]
 \centering
 \scriptsize
 \caption{Uncertainty-aware calibration ablation experiments on the Epinions dataset.}
 \label{tab:ablation_calibration}
 \resizebox{\columnwidth}{!}{%
 \begin{tabular}{@{}lcccc@{}}
 \toprule
 Model& ECE $\downarrow$ & Brier $\downarrow$ & NLL $\downarrow$ & Effect\\
 \midrule
 Full \model{} & 0.0158 & 0.0272 & 0.1084 & Complete reliability modeling\\
 w/o Uncertainty Calibration & 0.0185 & 0.0609 & 0.2022 & Remove uncertainty calibration loss\\
 w/o Post-hoc Calibration & 0.2647 & 0.2280 & 0.5872 & Remove validation set logit calibration\\
 \bottomrule
 \end{tabular}%
 }
\end{table}

\begin{figure}[!htbp]
 \centering
 \includegraphics[width=0.96\columnwidth]{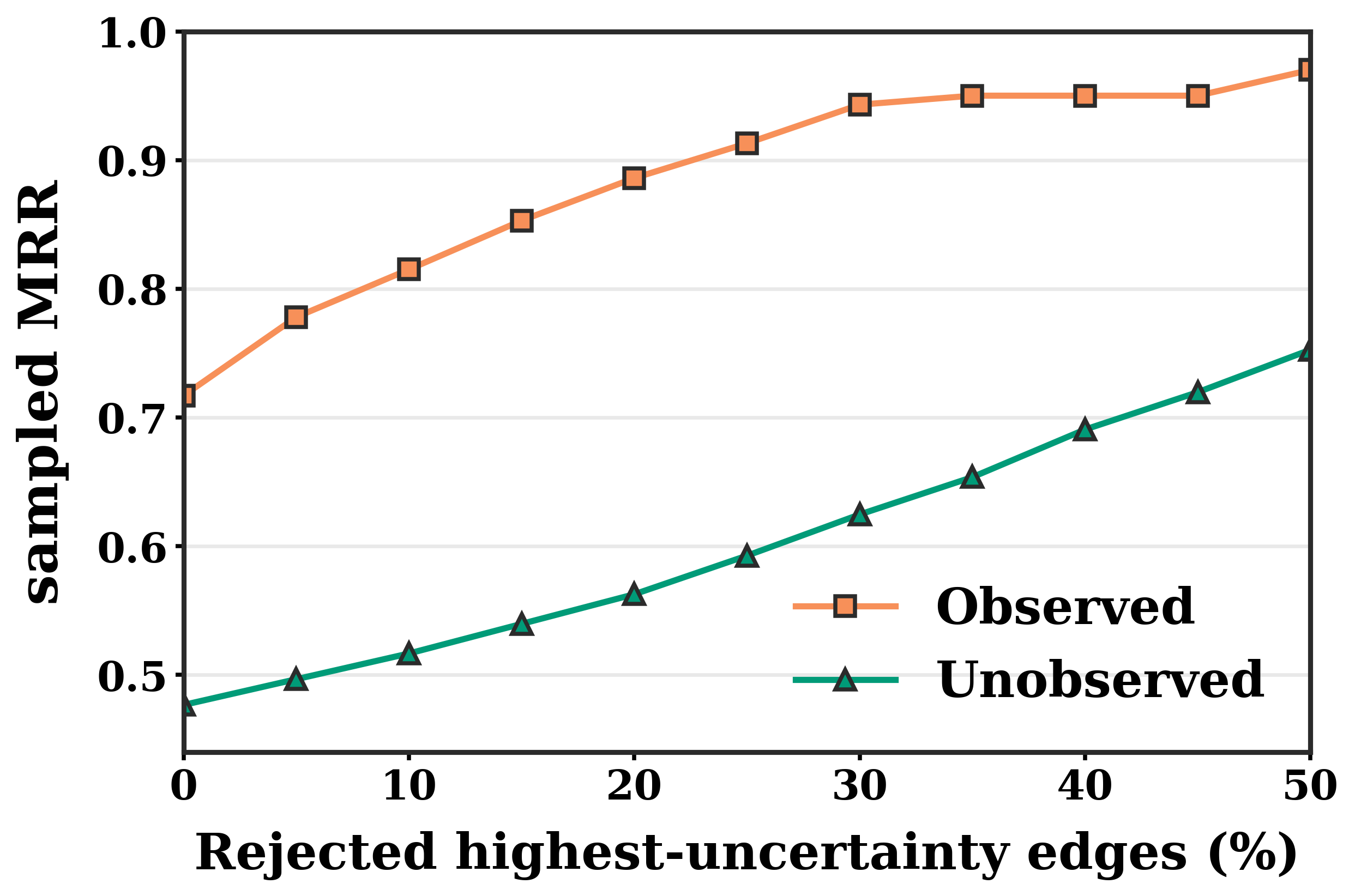}
 \caption{Selective prediction results under the partitioning of Epinions dataset 70\%-15\%-15\%. The horizontal axis represents the proportion of candidate edges rejected from high to low prediction uncertainty, and the vertical axis represents the sampled MRR on the held-out sample.}
 \label{fig:selective_prediction}
\end{figure}

Figure~\ref{fig:selective_prediction} tests whether uncertainty scores identify low-reliability predictions. Under the Epinions 70\%-15\%-15\% split, candidate edges are rejected from high to low uncertainty, and sampled MRR is recomputed on retained samples. The no-rejection points align with the Full \model{} MRR values in Table~\ref{tab:overall_epinions}.

After rejecting the top 50\% most uncertain edges, sampled MRR rises from 0.7171 to 0.9701 for observed users and from 0.4767 to 0.7527 for unobserved users. This shows that uncertainty scores align with error risk and can support reliability screening in cold-start or history-sparse cases.

Overall, the ablation experiments separate the roles of the main modules: conditioned propagation is the primary source of ranking gains, temporal memory improves historical-state representation, and calibration improves probability quality. The full model therefore cannot be explained by simple feature injection or posterior calibration alone.

\subsection{Hyperparameter Sensitivity Analysis}

To examine whether the main gains depend on a narrow hyperparameter choice, we further test the sensitivity of the candidate-operator number $K$ and the component-specific temporal decay coefficients $(\lambda_E,\lambda_B,\lambda_C)$ on Epinions under the 70\%-15\%-15\% split. For compactness, this subsection reports overall test-set MRR.

\begin{figure}[!htbp]
 \centering
 \includegraphics[width=0.82\columnwidth]{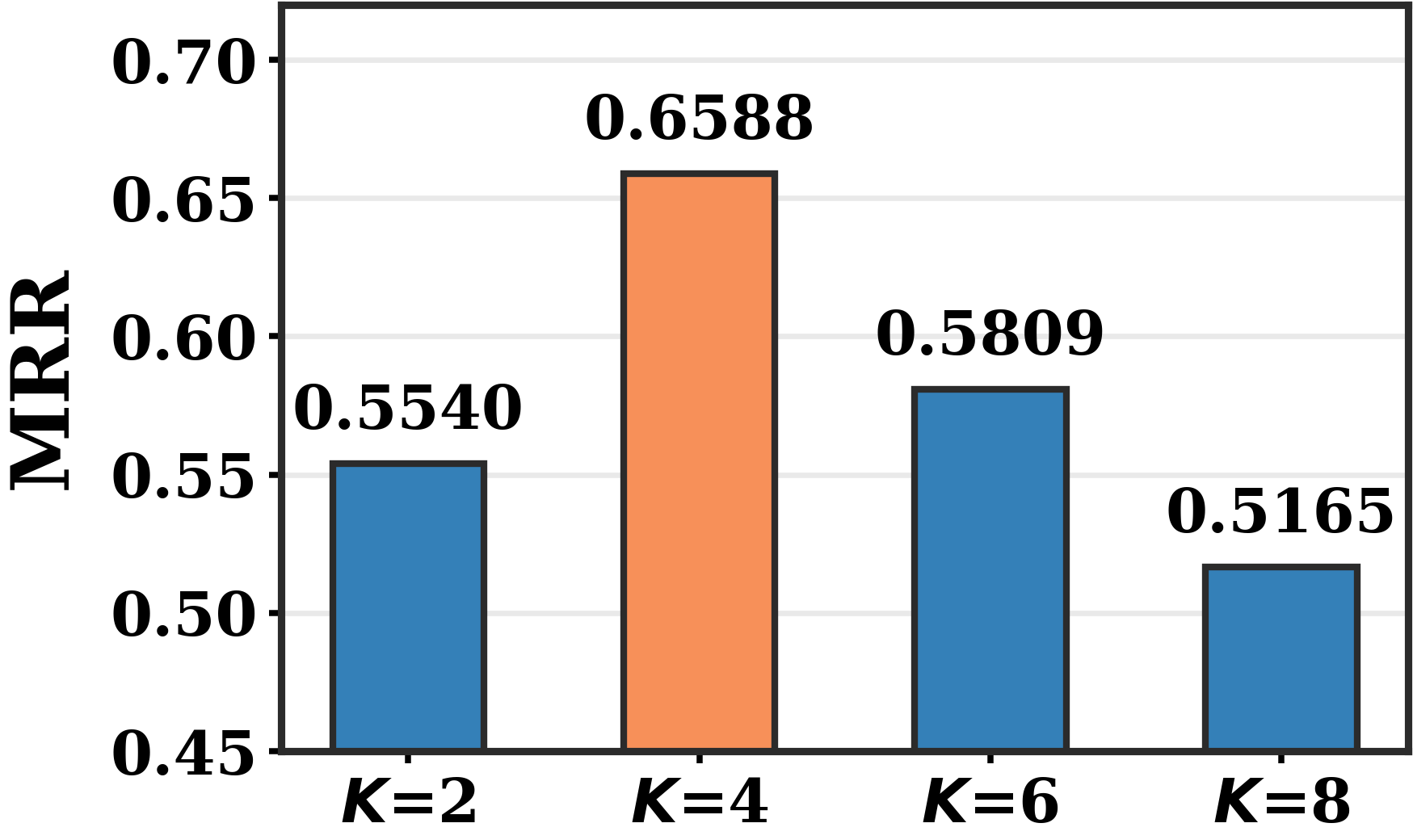}
 \vspace{-4pt}
 \caption{Sensitivity to the number of candidate heterogeneous propagation operators $K$ on Epinions under the 70\%-15\%-15\% split. The vertical axis reports overall test-set MRR.}
 \label{fig:k_sensitivity}
\end{figure}

Figure~\ref{fig:k_sensitivity} evaluates $K\in\{2,4,6,8\}$. The best result is obtained at $K=4$ with MRR $0.6588$, while reducing the candidate set to $K=2$ lowers MRR to $0.5540$. Increasing the operator bank beyond this point is also harmful: MRR drops to $0.5809$ at $K=6$ and to $0.5165$ at $K=8$. This pattern shows that context-conditioned propagation does not benefit from arbitrarily large operator banks. Too few candidates restrict propagation-mode expressiveness, whereas too many candidates introduce redundant choices and weaken stable operator selection.

\begin{figure}[!htbp]
 \centering
 \includegraphics[width=\columnwidth]{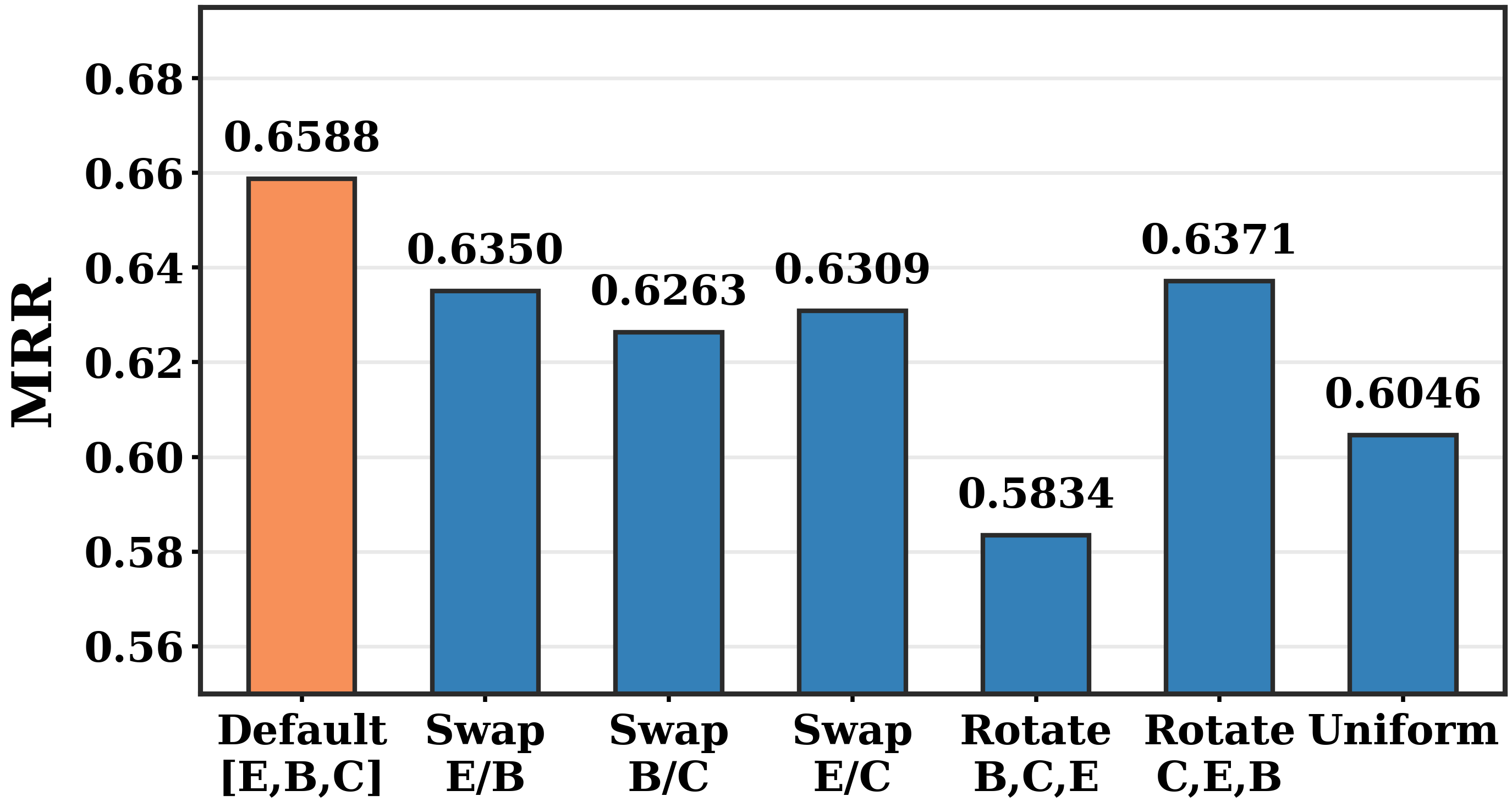}
 \vspace{-9pt}
 \caption{Sensitivity to component-specific temporal decay assignments on Epinions under the 70\%-15\%-15\% split. The vertical axis reports overall test-set MRR.}
 \label{fig:lambda_sensitivity}
\end{figure}

Figure~\ref{fig:lambda_sensitivity} evaluates the temporal-decay design. The default non-uniform setting $(\lambda_E,\lambda_B,\lambda_C)=(0.015,0.045,0.090)$ achieves the highest MRR among the main reference settings. Replacing it with uniform decay reduces MRR to $0.6046$. Permuting the three decay rates also generally weakens performance: swapping $\lambda_B$ and $\lambda_C$ yields $0.6263$, swapping $\lambda_E$ and $\lambda_C$ yields $0.6309$, and the cyclic permutation $[\lambda_B,\lambda_C,\lambda_E]$ drops further to $0.5834$. Two reordered cases remain closer to the default setting, namely swapping $\lambda_E$ and $\lambda_B$ ($0.6350$) and the cyclic permutation $[\lambda_C,\lambda_E,\lambda_B]$ ($0.6371$), but they still do not surpass the default configuration. Overall, these results support the claim that \model{} benefits not only from non-uniform decay itself, but also from assigning evidence-specific temporal scales in a semantically aligned way.

\subsection{Temporal Trust Memory Analysis}

\subsubsection{Predictive Performance for Medium Time-Interval Samples}

We divide test samples into short-, medium-, and long-interval groups according to the time gap $\Delta t$ between a target candidate edge and its most recent related historical interaction. Medium-interval samples are those between the 33rd and 67th percentiles, reducing interference from extremely short recency effects and extreme long-gap sparsity.

Table~\ref{tab:medium_gap} reports memory ablations on medium-interval samples. Full \model{} achieves 0.6292 MRR, higher than 0.5400 for w/o Masked Memory and 0.4868 for Uniform Decay Memory; AUC, AP, F1, and Accuracy are also best. The weaker Uniform Decay Memory result shows that sharing one decay mechanism across entity, behavior, and context evidence weakens historical-state representation. Temporal memory is useful because it preserves dynamic states for different evidence types, not because it adds a generic temporal encoding.

\begin{table}[!htbp]
 \centering
 \small
 \caption{Prediction performance on medium time-interval samples in the Epinions dataset.}
 \label{tab:medium_gap}
 \resizebox{\columnwidth}{!}{%
 \begin{tabular}{@{}lccccc@{}}
 \toprule
 Model& MRR & AUC & AP & F1 & Accuracy\\
 \midrule
 Full \model{} & 0.6292 & 0.9854 & 0.9803 & 0.9599 & 0.9584\\
 w/o Masked Memory & 0.5400 & 0.9749 & 0.9652 & 0.9461 & 0.9426\\
 Uniform Decay Memory & 0.4868 & 0.9733 & 0.9586 & 0.9439 & 0.9405\\
 \bottomrule
 \end{tabular}%
 }
\end{table}

\subsubsection{Statistical Differences Between Three Memory States}

\begin{figure}[!htbp]
 \centering
 \includegraphics[width=\columnwidth]{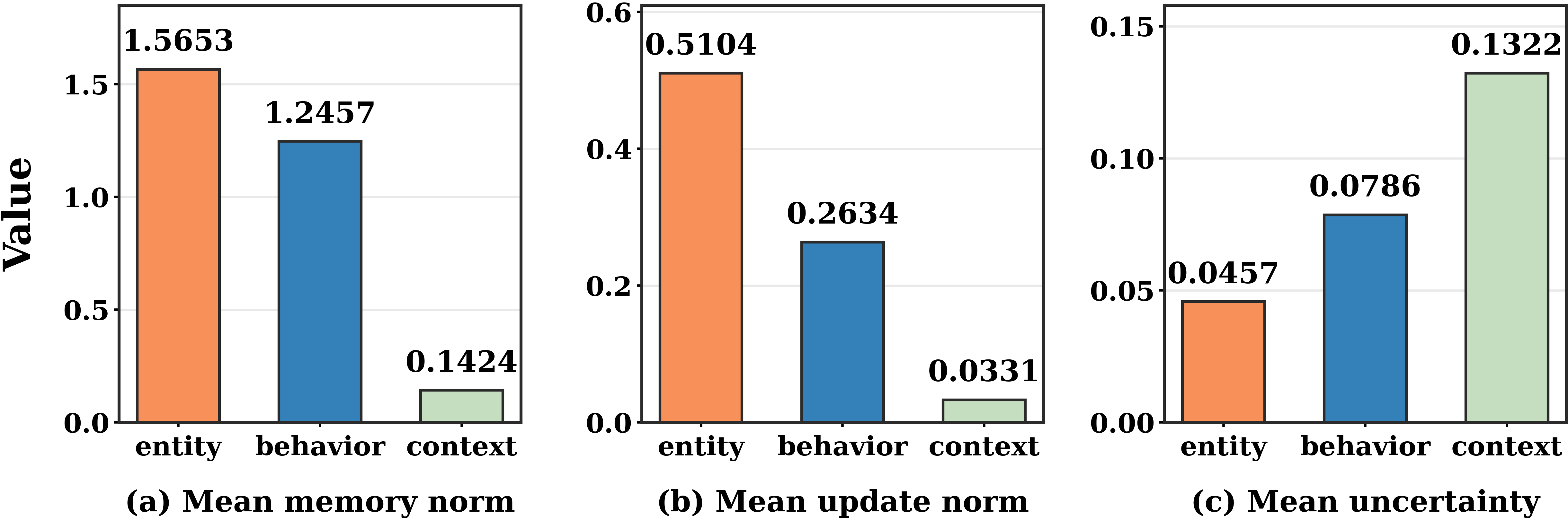}
 \vspace{-4pt}
 \caption{Entity, behavior, and contextual memory state statistics. Subfigures (a) -- (c) show the average memory norm, average update norm and average uncertainty, respectively.}
 \label{fig:memory_stats}
\end{figure}

Table~\ref{tab:memory_stats} and Figure~\ref{fig:memory_stats} show the statistics of the three memory states. Entity memory has the largest norm (1.5653) and the lowest uncertainty (0.0457), matching long-term reliability accumulation. Behavior memory is intermediate, reflecting both historical interaction patterns and recent behavioral changes. Context memory has the smallest norm (0.1424) but the highest uncertainty (0.1322), consistent with the expectation in Section 4.4: context mainly affects current propagation through operator selection, while memory keeps only weak historical residuals. These differences support component-wise memory and non-uniform decay.

\begin{table}[!htbp]
 \centering
 \small
 \caption{Component memory trace statistics.}
 \label{tab:memory_stats}
 \resizebox{\columnwidth}{!}{%
 \begin{tabular}{@{}lccc@{}}
 \toprule
 Component& Average memory norm& Average update norm& Average uncertainty\\
 \midrule
 Entity& 1.5653 & 0.5104 & 0.0457\\
 Behavior& 1.2457 & 0.2634 & 0.0786\\
 Context& 0.1424 & 0.0331 & 0.1322\\
 \bottomrule
 \end{tabular}%
 }
\end{table}

\subsection{Efficiency Analysis}

Table~\ref{tab:efficiency} compares the single-epoch training time, trainable parameters, and peak GPU memory of Linear, CAT, HGT, and \model{} under the Epinions 70\%-15\%-15\% split. \model{} is not the smallest model, but it trains faster and uses much less memory than CAT. This difference mainly comes from their computation patterns. CAT explicitly models context through context-aware meta-paths, context embeddings, and dual attention over sampled heterogeneous neighbors, which generates many intermediate graph messages and context-related representations. Therefore, its cost is dominated by graph-level path modeling and neighbor-level attention rather than parameter size. In contrast, \model{} converts entity reliability, interaction-behavior reliability, and contextual trust into compact propagation-control signals. Message admission and behavior modulation are implemented as lightweight gates or coefficients, while contextual trust selects among a small set of propagation operators instead of expanding context-specific paths. Although \model{} has more parameters than CAT, these modules operate on low-dimensional evidence states, leading to lower runtime and memory cost. Compared with HGT, \model{} uses fewer parameters and less GPU memory while achieving stronger predictive performance. These results indicate that \model{}'s gain does not come from a larger model scale, but from efficient evidence-controlled propagation.

\begin{table}[!htbp]
 \centering
 \small
 \caption{Efficiency comparison on Epinions under the 70\%-15\%-15\% split.}
 \label{tab:efficiency}
 \resizebox{\columnwidth}{!}{%
 \begin{tabular}{@{}lccc@{}}
 \toprule
 Model& Single-epoch training time (s)& Trainable parameters& GPU memory (MB)\\
 \midrule
 Linear & 0.4050 & 95 & 43.6000\\
 CAT & 57.6880 & 41.4K & 2699.2000\\
 HGT & 1.5910 & 677.0K & 916.3000\\
 \model{} & 14.7330 & 111.6K & 649.3000\\
 \bottomrule
 \end{tabular}%
 }
\end{table}

\subsection{Robustness Analysis}

To align with CAT's robustness experimental design, we conduct CAT-style training-stage poisoning experiments on Epinions under the 70\%-15\%-15\% and 80\%-10\%-10\% temporal splits. Two attack types are considered. The first is a trust-oriented attack, which performs local node-level trust manipulation: malicious users modify trust relations associated with target users through collusive bad-mouthing, thereby damaging the trustworthiness of target nodes. The second is a GNN-oriented attack, which performs global graph-structure poisoning: the attacker observes the contextual trust network, its statistics, and its temporal evolution, and perturbs the graph by adding or removing links. Following CAT, $p$ denotes the ratio of poisoned links to original links, and we evaluate $p=5\%,10\%,15\%,20\%$. MDR in Table~\ref{tab:pollution_robustness_table} denotes the maximum relative drop from the corresponding Clean MRR. Lower MDR indicates smaller relative degradation, while higher attacked MRR indicates stronger absolute performance under poisoning.

\begin{figure*}[!t]
 \centering
 \includegraphics[width=0.82\textwidth]{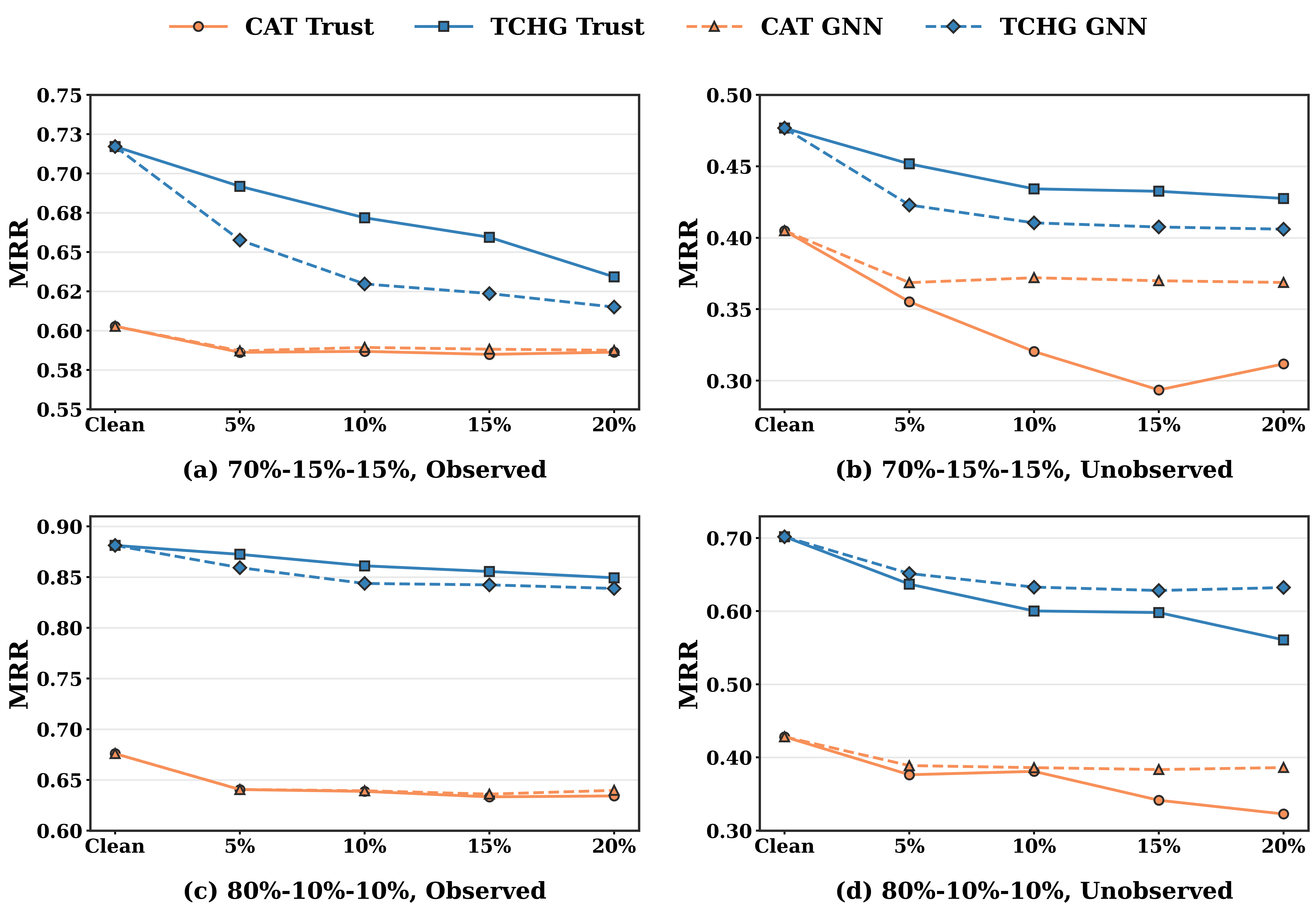}
 \caption{CAT-style robustness comparison on Epinions. Solid lines denote trust-oriented attacks and dashed lines denote GNN-oriented attacks; subfigures report observed-user and unobserved-user MRR under the 70\%-15\%-15\% and 80\%-10\%-10\% splits.}
 \label{fig:pollution_robustness}
\end{figure*}

Table~\ref{tab:pollution_robustness_table} and Figure~\ref{fig:pollution_robustness} show three findings. First, across all non-zero perturbation settings, \model{} obtains higher attacked MRR than CAT, indicating that its advantage is not limited to clean data. Second, relative degradation is attack- and split-dependent. Under the 70\%-15\%-15\% observed-user setting, \model{} suffers larger relative drops than CAT, although its attacked MRR remains higher. Under the 80\%-10\%-10\% observed-user setting, \model{} shows both higher attacked MRR and lower MDR. This suggests that the conditioned propagation and component memory become more stable when richer historical evidence is available. Third, unobserved-user prediction remains the harder robustness case. Trust-oriented attacks directly contaminate trust evidence and have a particularly strong impact when historical evidence is sparse, whereas GNN-oriented attacks perturb temporal neighborhoods and relation structures used by graph propagation. Therefore, the robustness results should be interpreted as improved absolute performance under CAT-style poisoning attacks, rather than as uniformly smaller relative degradation under all attack settings.

\begin{table*}[!t]
 \centering
 \scriptsize
 \caption{CAT-style robustness comparison on Epinions. Bold indicates the better result between CAT and \model{} for each task, split, attack, and metric. Higher MRR and lower MDR indicate better performance.}
 \label{tab:pollution_robustness_table}
 \resizebox{\textwidth}{!}{%
 \begin{tabular}{@{}lllccccccccccc@{}}
 \toprule
 Split & Task & Model & Clean & \multicolumn{5}{c}{Trust-oriented attack} & \multicolumn{5}{c}{GNN-oriented attack}\\
 \cmidrule(lr){5-9}\cmidrule(l){10-14}
 & & & MRR & $p=5\%$ & $p=10\%$ & $p=15\%$ & $p=20\%$ & MDR$\downarrow$ & $p=5\%$ & $p=10\%$ & $p=15\%$ & $p=20\%$ & MDR$\downarrow$\\
 \midrule
 \multirow{4}{*}{70\%-15\%-15\%} & \multirow{2}{*}{Observed} & CAT & 0.6028 & 0.5862 & 0.5868 & 0.5849 & 0.5863 & \bestval{2.97\%} & 0.5871 & 0.5893 & 0.5882 & 0.5875 & \bestval{2.60\%}\\
 & & \model{} & \bestval{0.7171} & \bestval{0.6918} & \bestval{0.6718} & \bestval{0.6594} & \bestval{0.6342} & 11.56\% & \bestval{0.6576} & \bestval{0.6297} & \bestval{0.6236} & \bestval{0.6150} & 14.24\%\\
 \cmidrule(lr){2-14}
 & \multirow{2}{*}{Unobserved} & CAT & 0.4049 & 0.3552 & 0.3204 & 0.2935 & 0.3117 & 27.51\% & 0.3686 & 0.3720 & 0.3699 & 0.3687 & \bestval{8.97\%}\\
 & & \model{} & \bestval{0.4767} & \bestval{0.4517} & \bestval{0.4342} & \bestval{0.4325} & \bestval{0.4275} & \bestval{10.32\%} & \bestval{0.4229} & \bestval{0.4104} & \bestval{0.4075} & \bestval{0.4060} & 14.83\%\\
 \midrule
 \multirow{4}{*}{80\%-10\%-10\%} & \multirow{2}{*}{Observed} & CAT & 0.6758 & 0.6404 & 0.6386 & 0.6332 & 0.6341 & 6.30\% & 0.6405 & 0.6392 & 0.6359 & 0.6396 & 5.90\%\\
 & & \model{} & \bestval{0.8811} & \bestval{0.8723} & \bestval{0.8610} & \bestval{0.8554} & \bestval{0.8492} & \bestval{3.62\%} & \bestval{0.8592} & \bestval{0.8436} & \bestval{0.8422} & \bestval{0.8387} & \bestval{4.81\%}\\
 \cmidrule(lr){2-14}
 & \multirow{2}{*}{Unobserved} & CAT & 0.4282 & 0.3763 & 0.3809 & 0.3414 & 0.3226 & 24.66\% & 0.3888 & 0.3860 & 0.3834 & 0.3861 & \bestval{10.46\%}\\
 & & \model{} & \bestval{0.7018} & \bestval{0.6370} & \bestval{0.6003} & \bestval{0.5981} & \bestval{0.5608} & \bestval{20.09\%} & \bestval{0.6515} & \bestval{0.6329} & \bestval{0.6283} & \bestval{0.6323} & 10.47\%\\
 \bottomrule
 \end{tabular}%
 }
\end{table*}

\section{Discussion}

This section discusses the main implications of the experimental results and clarifies the scope of the conclusions. We focus on three issues: how evidence-controlled propagation differs from feature enrichment, how the results should be interpreted under different timestamp settings, and what kind of reliability and robustness \model{} provides.

\subsection{Evidence-Controlled Propagation}

The results suggest that the MRR advantage of \model{} mainly comes from evidence-controlled propagation, rather than from feature enrichment alone. Both Feature Injection Only and Attention Only produce lower MRR than the full model, indicating that tri-trust evidence should not be used only as additional input features or generic attention signals. In \model{}, entity reliability, interaction-behavior reliability, and contextual trust are mapped to message admission, propagation-strength modulation, and operator selection, respectively. This allows different types of evidence to play distinguishable roles during graph propagation.

The temporal-memory results further support this design. Entity evidence tends to preserve long-term reliability states, behavior evidence captures more recent interaction changes, and contextual evidence is more volatile. Therefore, compared with a unified temporal state, component-decoupled memory with non-uniform decay is more suitable for modeling the asynchronous evolution of trust evidence.

\subsection{Dataset Scope and Dynamic Modeling}

The empirical conclusions of this paper should be interpreted according to the timestamp availability of each dataset. Epinions provides native trust timestamps and is therefore used as the main dynamic trust prediction setting, with separate evaluation for observed-user and unobserved-user scenarios. Ciao and CiaoDVD do not provide native trust timestamps. We therefore evaluate them under all-users random trust-link prediction splits without constructing pseudo trust timestamps.

Accordingly, the dynamic modeling claim is mainly supported by Epinions, where trust events can be ordered by native timestamps. The results on Ciao and CiaoDVD show that tri-trust evidence modeling remains effective when native trust timestamps are unavailable, and further demonstrate its general applicability to all-users trust-link prediction.

\subsection{Reliability and Robustness}

\model{} improves not only ranking performance but also the reliability of probability outputs. Ranking metrics, including MRR, AP, and AUC, are computed from pre-calibration outputs, whereas reliability metrics such as ECE, Brier Score, and NLL are computed from calibrated probabilities. Thus, calibration is used to support reliable decision making in risk-sensitive scenarios, rather than to improve ranking results.

The robustness results should be interpreted in the same cautious manner. Under trust-oriented and GNN-oriented poisoning attacks, \model{} generally maintains higher attacked MRR than CAT, suggesting that its advantage is not limited to clean data. In this sense, \model{} achieves stronger absolute performance under the tested attacks. However, the relative degradation still depends on the attack type and data split, and there remains room for improvement in some settings.

\section{Conclusion}

In this paper, we propose \model{}, which organizes trust evidence into entity, behavior, and context channels, and integrates tri-trust conditioned propagation, component-decoupled temporal memory, and uncertainty-aware output modeling into a unified framework. In this way, \model{} not only captures heterogeneous trust dependencies, but also models asynchronous evidence evolution and improves the reliability of probability outputs. Experiments on Epinions, Ciao, and CiaoDVD demonstrate the effectiveness of \model{} against representative trust prediction and heterogeneous graph baselines, with ablation, memory, efficiency, robustness, and selective prediction analyses further supporting the proposed designs. The current framework still has several limitations. Its evidence construction mainly relies on trust links, ratings, temporal statistics, and local graph context, while richer signals such as review text, user profiles, item semantics, and multimodal behavior have not yet been incorporated. Future work will extend \model{} with richer platform evidence, more complete temporal event streams, and stronger evaluation under adaptive and collusive attacks.

\section*{Acknowledgment}

This work was supported by the Special Project of the National Natural Science Foundation of China (Grant No. 62341128) and the Natural Science Foundation of Shaanxi Province (Grant No. 2025JC-YBMS-786).

\end{document}